\documentclass[11pt]{article}

\usepackage[final]{acl}

\usepackage{booktabs}
\usepackage{multirow}
\usepackage{times}
\usepackage{latexsym}
\usepackage{amssymb}
\usepackage[T1]{fontenc}
\usepackage{amsthm}
\usepackage[utf8]{inputenc}
\usepackage{amsmath}
\usepackage{microtype}
\usepackage{amsthm}

\usepackage{inconsolata}

\usepackage{graphicx}
\usepackage{colortbl}
\usepackage[table]{xcolor}
\usepackage{amsmath}
\usepackage{subcaption}
\usepackage{enumitem}
\usepackage{xurl}
\usepackage[table]{xcolor}
\usepackage{booktabs}
\usepackage{array} 

\definecolor{lightblue}{HTML}{E6F0FF} 
\definecolor{green}{RGB}{0, 120, 0} 

\title{Outcome-Grounded Advantage Reshaping for Fine-Grained Credit Assignment in Mathematical Reasoning}

\author{
 \textbf{Ziheng Li\textsuperscript{1,2,3}},
 \textbf{Liu Kang\textsuperscript{2}},
 \textbf{Feng Xiao\textsuperscript{2}},
 \textbf{Luxi Xing\textsuperscript{2}},
 \textbf{Qingyi Si\textsuperscript{4}},
  \textbf{Zhuoran Li\textsuperscript{5}},
  \\
  \textbf{Weikang Gong\textsuperscript{1}},
  \textbf{Deqing Yang\textsuperscript{1}},
 \textbf{YangHua Xiao\textsuperscript{1}},
 \textbf{Hongcheng Guo\textsuperscript{1}\thanks{Corresponding author.}}
\\
\\
 \textsuperscript{1}Fudan University
 \textsuperscript{2}XingYun lab, HUJING Digital Media \& Entertainment Group
\\
\textsuperscript{3}University of Science and Technology Beijing 
 \textsuperscript{4}Chinese Academy of Sciences \\
  \textsuperscript{5}Beijing University of Posts and Telecommunications 
}
\usepackage{amsthm}

\begin{document}
\maketitle
\begin{abstract}
Group Relative Policy Optimization (GRPO) has emerged as a promising critic-free reinforcement learning paradigm for reasoning tasks. 
However, standard GRPO employs a coarse-grained credit assignment mechanism that propagates group-level rewards uniformly to to every token in a sequence, neglecting the varying contribution of individual reasoning steps. 
We address this limitation by introducing \textbf{O}utcome-grounded \textbf{A}dvantage \textbf{R}eshaping (\textbf{OAR}), a fine-grained credit assignment mechanism that redistributes advantages based on how much each token influences the model's final answer.
We instantiate OAR via two complementary strategies: 
(1) \textbf{OAR-P}, which estimates outcome sensitivity through counterfactual token perturbations, serving as a high-fidelity attribution signal;
(2) \textbf{OAR-G}, which uses an input-gradient sensitivity proxy to approximate the influence signal with a single backward pass. 
These importance signals are integrated with a conservative \textbf{Bi-Level} advantage reshaping scheme that suppresses low-impact tokens and boosts pivotal ones while preserving the overall advantage mass. 
Empirical results on extensive mathematical reasoning benchmarks demonstrate that while OAR-P sets the performance upper bound, OAR-G achieves comparable gains with negligible computational overhead, both significantly outperforming a strong GRPO baseline, pushing the boundaries of critic-free LLM reasoning~\footnote{code at: \url{https://github.com/XINGYUN-AI-LAB/OAR}}.

\end{abstract}

\section{Introduction}
Large Language Models (LLMs) have recently demonstrated strong performance on complex reasoning tasks \citep{fedus2022switchtransformersscalingtrillion, achiam2023gpt4,brown2020languagemodelsfewshotlearners}, driven in part by reinforcement learning with verifiable rewards \citep[RLVR;][]{lambert2024tulu}. Among these approaches, Group Relative Policy Optimization \citep[GRPO;][]{Shao2024DeepSeekMathPT} has emerged as a practical critic-free algorithm for post-training reasoning models. By normalizing rewards over a group of sampled responses, GRPO avoids training a value function while remaining stable and scalable for long-horizon reasoning.

\begin{figure}[t]
    \centering
    \includegraphics[width=\columnwidth]{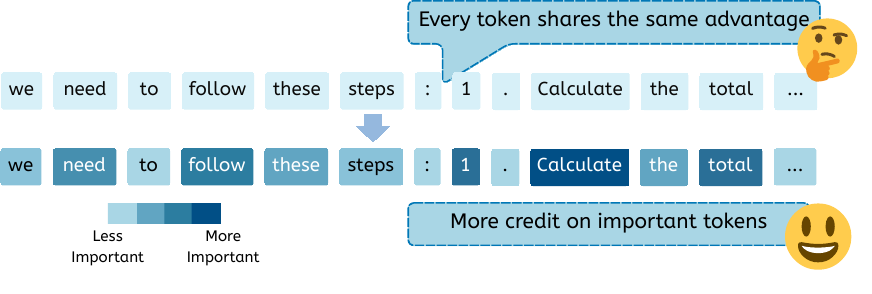}
    \caption{From broadcast credit to token reallocation.}
    \label{fig:compare}
\end{figure}

However, GRPO relies on a coarse credit assignment: a single group-normalized advantage is broadcast to all tokens in a sampled response \cite{wei2023chainofthoughtpromptingelicitsreasoning}. This ignores the uneven structure of reasoning traces, where only a small set of tokens (e.g., key deductions or structural decisions) determines correctness, while many others are largely syntactic \cite{yang2025letlowprobabilitytokensoverdominate}.  Figure~\ref{fig:compare} illustrates this mismatch between coarse sequence-level credit assignment and the desired token-level reallocation. Treating all positions equally increases gradient variance and can accelerate entropy collapse during training \citep{liu2024tis,li2024preserving}.

A natural direction is to move from coarse sequence-level credit to token-level credit. Recent work uses token entropy as a proxy to reweight updates: \citet{cheng2025reasoningexplorationentropyperspective} amplifies high-entropy tokens to encourage exploration, while \citet{chen2025highentropyexplorationcorrectnessawarelowentropy} shapes advantages over low-entropy segments in a correctness-aware way to consolidate useful structures and suppress recurring failure patterns. However, entropy primarily reflects uncertainty or stability\cite{pmlr-v48-gal16}, not outcome-relevant importance, and may therefore misallocate credit to tokens that are salient to the policy yet only weakly influential on the final answer outcome\cite{jain2019attentionexplanation}.

In contrast, we argue that the ideal token-level credit in critic-free RL should be outcome-grounded, reflecting how much each token influences the model’s final answer. To this end, we propose \textbf{O}utcome-grounded \textbf{A}dvantage \textbf{R}eshaping (\textbf{OAR}), a framework that enhances GRPO with outcome-sensitive token attribution. OAR scores each token by how much perturbing it shifts the model's final-answer distribution, inspired by perturbation-based feature attribution\cite{ribeiro2016whyitrustyou}. While the ideal attribution would measure reward changes under interventions, RLVR rewards are typically rule-based, sparse, and non-differentiable, making token--reward attribution prohibitively expensive. We therefore use the model's own answer distribution as a practical surrogate outcome to estimate token impact.

We instantiate OAR with two complementary attribution strategies that trade off fidelity and efficiency: \textbf{OAR-P} estimates outcome sensitivity via counterfactual token perturbations, providing a high-fidelity but costly signal, while \textbf{OAR-G} uses an efficient gradient-based proxy suitable for scalable online training\cite{simonyan2014deepinsideconvolutionalnetworks}. However, attribution scores alone do not specify \emph{how} to inject token-level credit without destabilizing GRPO: naively multiplying or adding token weights can induce sample-dependent changes in the effective update scale.

Therefore, we introduce a conservative \textbf{Bi-Level} advantage reallocation mechanism that suppresses low-impact tokens to reduce gradient noise, boosts high-impact tokens to strengthen the learning signal, and renormalizes weights to preserve the overall advantage mass.


Our contributions can be summarized as follows:
\begin{itemize}[itemsep=5pt, parsep=0pt, topsep=0pt, partopsep=0pt]
    \item We propose \textbf{OAR}, an outcome-grounded framework for fine-grained credit assignment, which redistributes sequence-level advantages based on token influence on the model's final answer distribution, with two instantiations: \textbf{OAR-P} (perturbation-based attribution) and \textbf{OAR-G} (gradient-based approximation).
    \item We design a conservative \textbf{Bi-Level} advantage reallocation mechanism that sharpens learning signals by boosting pivotal tokens and suppressing low-influence noise, while preserving the overall advantage mass to maintain stable training.
    \item Extensive experiments on mathematical reasoning benchmarks demonstrate that OAR consistently outperforms strong GRPO baselines. While OAR-P sets the performance upper bound, OAR-G retains the majority of these gains with minimal computational cost.
\end{itemize}

\section{Preliminaries}

\subsection{Group Relative Policy Optimization}

A large language model defines an autoregressive policy $\pi_\theta$ that generates a token sequence $y=(y_1,\dots,y_T)$.  
In RLVR, we maximize the expected sequence-level reward:
\begin{equation}
    J(\theta) = \mathbb{E}_{x\sim\mathcal{D},\, y\sim\pi_\theta(\cdot|x)}\big[ r(y) \big].
\end{equation}

GRPO eliminates the learned value function in PPO by using a group-normalized outcome reward as the advantage.  
Given a prompt $x$, we sample $G$ completions $\{y^{(1)},\dots,y^{(G)}\}$ from the behavior policy $\pi_{\theta_{\text{old}}}$ and obtain rewards $\{r_1,\dots,r_G\}$. The advantage for sample $i$ is computed by within-group normalization:
\begin{equation}
\hat{A}_i
= \frac{r_i - \frac{1}{G}\sum_{k=1}^{G} r_k}
{\sqrt{\frac{1}{G}\sum_{k=1}^{G}\left(r_k - \frac{1}{G}\sum_{j=1}^{G} r_j\right)^2}}.
\end{equation}
This scalar $\hat{A}_i$ is applied to all token-level log-probability terms in $y^{(i)}$.

GRPO then optimizes the PPO-style clipped surrogate objective:
\begin{equation}
\label{eq:grpo_loss}
\begin{aligned}
\mathcal{L}_{\text{GRPO}}
= \mathbb{E}_{i,t}\Big[
\min\Big(
&\, \rho^{(i)}_t \hat{A}_i, \\
&\, \operatorname{clip}(\rho^{(i)}_t, 1-\epsilon, 1+\epsilon)\hat{A}_i
\Big)
\Big].
\end{aligned}
\end{equation}
with importance ratio
\begin{equation}
    \rho^{(i)}_t
    =
    \frac{\pi_\theta(y^{(i)}_t\,|\,x, y^{(i)}_{<t})}
         {\pi_{\theta_{\text{old}}}(y^{(i)}_t\,|\,x, y^{(i)}_{<t})}.
\end{equation}

While GRPO avoids value-function training, using a single sequence-level advantage for all tokens can increase gradient variance due to reward aliasing; see Appendix~\ref{subsubsec:variance_from_grpo}.

\begin{figure*}[t]
    \centering
    \includegraphics[width=\textwidth]{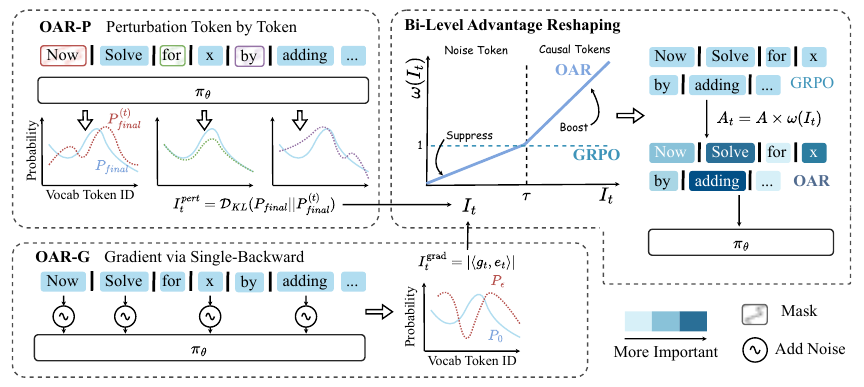} 
    \caption{Overall architecture of the proposed OAR framework integrated into GRPO.}
    \label{fig:architecture}
\end{figure*}

\subsection{Intrinsic Signals for Token-Level Credit}
\label{subsec:entropy_proxy}

Recent GRPO variants explore redistributing sequence-level advantages using intrinsic, model-derived signals as token-importance proxies \cite{li2025attention, xie2025unlockingexplorationrlvruncertaintyaware, chen2025highentropyexplorationcorrectnessawarelowentropy}. Among them, entropy-based advantage shaping \cite{cheng2025reasoningexplorationentropyperspective} is a representative and easily reproducible instance. For token distribution $\pi_\theta(\cdot \mid x, y_{<t})$, the token entropy is
\begin{equation}
H_t \;=\; -\sum_{v} \pi_\theta(v \mid x, y_{<t}) \log \pi_\theta(v \mid x, y_{<t}).
\label{eq:entropy}
\end{equation}
A typical shaping rule augments the broadcast advantage $A$ with an entropy-dependent bonus (Cheng et al., 2025):
\begin{equation}
\tilde{A}_t \;=\; A + \min\!\Big(\alpha \cdot \mathrm{sg}(H_t), \, |A|/\kappa\Big),
\label{eq:entropy_shaping}
\end{equation}
where $\mathrm{sg}(\cdot)$ stops gradients for stability.

While simple, intrinsic proxies may not reflect \emph{outcome-relevant} token importance: entropy measures predictive uncertainty and can be high for stylistic or lexically ambiguous tokens with little effect on correctness. Moreover, additive shaping alters the post-normalization advantage mass, inducing sequence-dependent update scales unless renormalization or re-tuning is applied (Theoretical Analysis is provided in Appendix~\ref{subsubsec:entropy_limitation}). 

\section{Method}
\label{sec:method}

We introduce \textbf{Outcome-grounded Advantage Reshaping (OAR)}, which equips GRPO with token-level credit assignment. OAR admits two instantiations: \textbf{OAR-P}, a perturbation-based attribution method, and \textbf{OAR-G}, an efficient gradient-based approximation.
Figure~\ref{fig:architecture} provides an overview of how OAR is integrated into GRPO.
\subsection{OAR-P}
\label{subsec:oar_p}
To establish a rigorous causal baseline, OAR-P employs a counterfactual masking strategy. Let $y = (y_1, \dots, y_T)$ be a reasoning chain generated by the policy $\pi_\theta$ given prompt $x$. We denote the model's final prediction distribution as $P_{\text{final}} = \pi_\theta(\cdot \mid x, y)$, which represents the probability distribution over the answer span. Here the answer span is extracted from the model output by a deterministic rule (e.g., regex matching \texttt{<answer>...\ </answer>}; see \ref{subsec:ablation_outcome_def}).

We construct a perturbed sequence $\tilde{y}^{(t)}$ by replacing the token at position $t$ with a special token (e.g., \texttt{[PAD]}):
\begin{equation}
    \tilde{y}^{(t)} = (y_1, \dots, y_{t-1}, \texttt{[PAD]}, y_{t+1}, \dots, y_T).
\end{equation}
Then perform a forward pass to compute the perturbed final distribution $P_{\text{final}}^{(t)} = \pi_\theta(\cdot \mid x, \tilde{y}^{(t)})$.

The causal importance score $I_t^{\text{pert}}$ for token $y_t$ is defined as the KL divergence between the original and the perturbed distributions:
\begin{equation}
\label{eq:importance_kl}
    I_t^{\text{pert}} = D_{\text{KL}}\left( P_{\text{final}} \,||\, P_{\text{final}}^{(t)} \right).
\end{equation}
Intuitively, $I_t^{\text{pert}}$ measures the information loss incurred by removing $y_t$. If $y_t$ is a logical pivot, masking it will significantly alter the final prediction ($I_t^{\text{pert}} \gg 0$); if $y_t$ is syntactic filler, the prediction remains stable ($I_t^{\text{pert}} \approx 0$). A thereotical perspective is provided in Appendix \ref{subsubsec:cfkl}. While accurate, calculating $I_t^{\text{pert}}$ for a sequence of length $L$ requires $L$ additional forward passes, creating a computational bottleneck.

\subsection{OAR-G}
\label{subsec:oar_g}
To enable scalable fine-grained credit assignment during online training, we propose OAR-G, which approximates the perturbation-based attribution using a single backward pass. Instead of discrete masking, we inject small Gaussian noise into the input representations of reasoning tokens and measure how sensitive the final outcome distribution is to each token.
Let $E=(e_1,\dots,e_T)$ denote the embeddings of the reasoning tokens in the sampled trajectory. We first compute the \emph{teacher} outcome distribution from the unperturbed input:
\begin{equation}
    P_{0} = P_{\text{final}}(x, E).
\end{equation}
Next, we apply isotropic Gaussian noise to the embeddings and obtain a \emph{student} outcome distribution under the perturbed representations:
\begin{equation}
    \tilde{e}_t = e_t + \epsilon_t, \epsilon_t \sim \mathcal{N}(0,\sigma^2 I),
    P_{\epsilon} = P_{\text{final}}(x, \tilde{E}).
\end{equation}
We then measure the induced distribution shift via a self-distillation objective:
\begin{equation}
\begin{split}
    \mathcal{J}(x, y) = D_{\text{KL}}\left(P_{0} \,||\, P_{\epsilon}\right)
    = \\ \sum_{v} P_{0}(v)\left(\log P_{0}(v) - \log P_{\epsilon}(v)\right).
\end{split}
\end{equation}
Token-wise sensitivity is obtained by differentiating $\mathcal{J}$ with respect to each input embedding:
\begin{equation}
    g_t = \nabla_{e_t} \mathcal{J}(x, y).
\end{equation}
Following gradient-based attribution methods~\citep{sundararajan2017axiomaticattributiondeepnetworks}, we define the importance score using \emph{Gradient $\times$ Input}:
\begin{equation}
\label{eq:importance_grad}
    I_t^{\text{grad}} = \left|\langle g_t, e_t \rangle\right|.
\end{equation}
This score corresponds to a first-order approximation of how strongly perturbing token $t$ would affect the outcome distribution. Importantly, OAR-G replaces $O(L)$ counterfactual forward passes with a single backward pass, making it computationally negligible compared to the generation process.

\paragraph{Normalization}
For both OAR-P and OAR-G, the raw importance scores ($I_t^{\text{pert}}$ or $I_t^{\text{grad}}$) are mapped to a unified scale for stability. We apply a log-transform followed by min-max normalization within each sequence:
\begin{equation}
    \bar{I}_t = \log(1 + I_t), \quad 
    \hat{I}_t = \frac{\bar{I}_t - \min_{j}\bar{I}_j}{\max_{j}\bar{I}_j - \min_{j}\bar{I}_j + \epsilon}.
\end{equation}
These normalized scores $\hat{I}_t$ are then used to modulate the advantage estimates in the subsequent reshaping phase.

\begin{table*}[t]
\centering
\small
\setlength{\tabcolsep}{7pt} 
\renewcommand{\arraystretch}{1}
\begin{tabular}{lcccccccccc}
\toprule
\multirow{2}{*}{Method} 
& \multicolumn{2}{c}{AIME25} 
& \multicolumn{2}{c}{AIME24} 
& \multicolumn{2}{c}{AMC23} 
& MATH500 & GSM8K & Avg \\
\cmidrule(lr){2-3} \cmidrule(lr){4-5} \cmidrule(lr){6-7} \cmidrule(lr){8-8} \cmidrule(lr){9-9} \cmidrule(lr){10-10}
& \textit{P@1} & \textit{P@32} 
& \textit{P@1} & \textit{P@32} 
& \textit{P@1} & \textit{P@32} 
& \textit{P@1} & \textit{P@1} & \textit{Avg} \\
\midrule
\textit{Qwen2.5-7B} & 2.5 & 26.4 & 4.5 & 43.6 & 29.5 & 81.4 & 52.7 & 82.0 & 45.2 \\
+ GRPO & 9.4 & 31.0 & 13.5 & 45.0 & 59.8 & 85.0 & 75.8 & 90.5 & 51.3 \\
+ GRPO w/ Random Adv & 9.2 & 31.0 & 13.2 & 44.3 & 60.1 & 85.5 & 75.2 & 90.3 & 51.1 \\
+ GRPO w/ Entropy Adv & 10.3 & 32.4 & 14.4 & 45.7 & \textbf{62.2} & 86.4 & 77.2 & 91.2 & 52.5 \\
\rowcolor{blue!5}  
+ GRPO w/ OAR-G & \underline{11.8} & \underline{33.2} & \underline{14.8} & \underline{47.8} & 61.4 & \underline{86.9} & \textbf{78.7} & \underline{91.4} & \underline{53.2} \\
\rowcolor{blue!10}  
+ GRPO w/ OAR-P & \textbf{12.2} & \textbf{33.9} & \textbf{15.2} & \textbf{48.2} & \underline{61.9} & \textbf{87.7} & \underline{78.4} & \textbf{92.0} & \textbf{53.7} \\
\midrule
$\Delta$ (vs. GRPO)
& \textcolor{green}{+2.8}
& \textcolor{green}{+2.9}
& \textcolor{green}{+1.7}
& \textcolor{green}{+3.2}
& \textcolor{green}{+2.1}
& \textcolor{green}{+2.7}
& \textcolor{green}{+2.6}
& \textcolor{green}{+1.5}
& \textcolor{green}{+2.4} \\
\midrule
\textit{Qwen2.5-Math-7B} & 5.7 & 33.5 & 11.8 & 49.4 & 33.2 & 88.3 & 51.2 & 69.0 & 42.8 \\
+ GRPO & 12.2 & 39.1 & 31.2 & 48.9 & 64.5 & 90.6 & 80.8 & 91.2 & 57.3 \\
+ GRPO w/ Random Adv & 11.9 & 39.4 & 30.8 & 47.2 & 64.0 & 90.2 & 79.8 & 91.2 & 56.8 \\
+ GRPO w/ Entropy Adv & 13.0 & 39.6 & 32.4 & 49.0 & \underline{65.4} & 90.6 & 81.7 & 91.4 & 57.9 \\
+ GRPO w/ KTAE & \textbf{15.2} & \textbf{42.4} & 33.3$^{\dagger}$ & 50.8 & 65.1$^{\dagger}$ & 89.7 & 82.4$^{\dagger}$ & \underline{92.7} & 59.0 \\
\rowcolor{blue!5}  
+ GRPO w/ OAR-G & 14.3 & 40.3 & \textbf{33.5} & \underline{51.7} & 65.2 & \underline{92.5} & \underline{83.0} & \textbf{92.9} & \underline{59.2} \\
\rowcolor{blue!10}  
+ GRPO w/ OAR-P & \underline{14.5} & \underline{40.8} & \textbf{33.5} & \textbf{51.9} & \textbf{65.6} & \textbf{92.8} & \textbf{83.8} & \underline{92.7} & \textbf{59.5} \\
\midrule
$\Delta$ (vs. GRPO)
& \textcolor{green}{+2.3} & \textcolor{green}{+1.7} & \textcolor{green}{+2.3} & \textcolor{green}{+3.0} & \textcolor{green}{+1.1} & \textcolor{green}{+2.2} & \textcolor{green}{+3.0} & \textcolor{green}{+1.5} & \textcolor{green}{+2.2} \\
\bottomrule
\end{tabular}
\caption{Main results on mathematical reasoning benchmarks. \textbf{bold} and \underline{underline} indicate the best and second-best results, respectively. $^{\dagger}$ indicates results quoted from the original KTAE~\cite{sun2025ktaemodelfreealgorithmkeytokens} report.}
\label{tab:main_results}
\end{table*}

\begin{figure*}[t]
\centering
\begin{minipage}{0.25\textwidth}\centering
  \quad \small Overall Reward
\end{minipage}\hfill
\begin{minipage}{0.25\textwidth}\centering
  \quad \small Accuracy Reward
\end{minipage}\hfill
\begin{minipage}{0.25\textwidth}\centering
  \quad \small Format Reward
\end{minipage}\hfill
\begin{minipage}{0.25\textwidth}\centering
  \quad \small Policy Entropy
\end{minipage}


\begin{minipage}{0.25\textwidth}\centering
  \includegraphics[width=\linewidth]{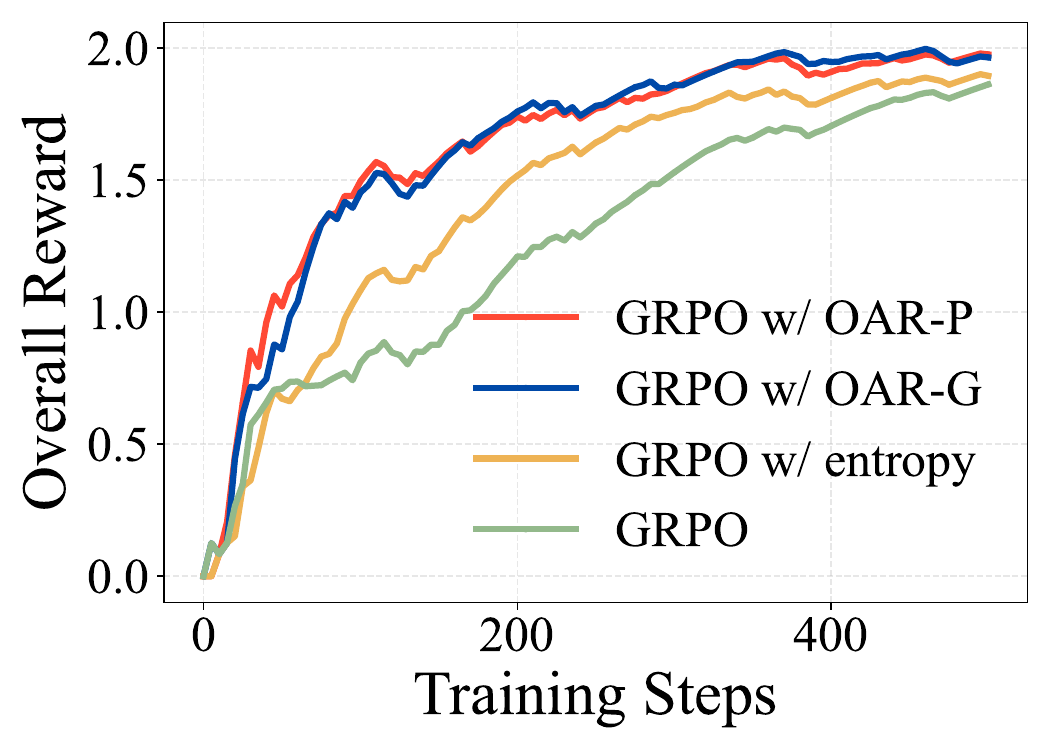}
\end{minipage}\hfill
\begin{minipage}{0.25\textwidth}\centering
  \includegraphics[width=\linewidth]{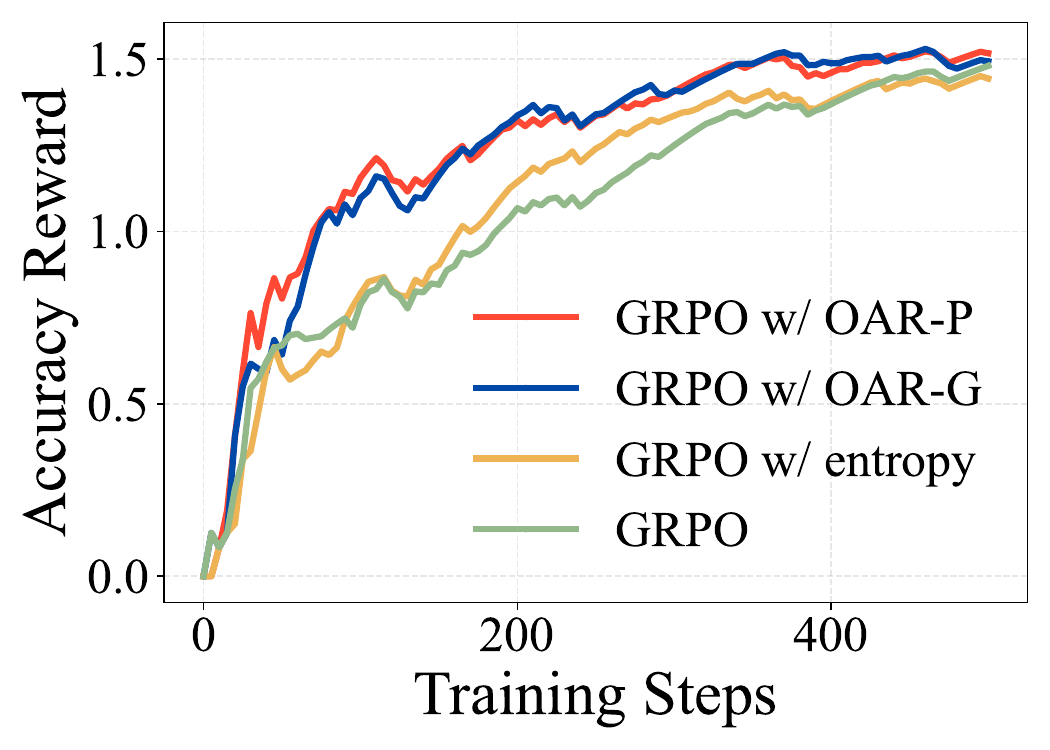}
\end{minipage}\hfill
\begin{minipage}{0.25\textwidth}\centering
  \includegraphics[width=\linewidth]{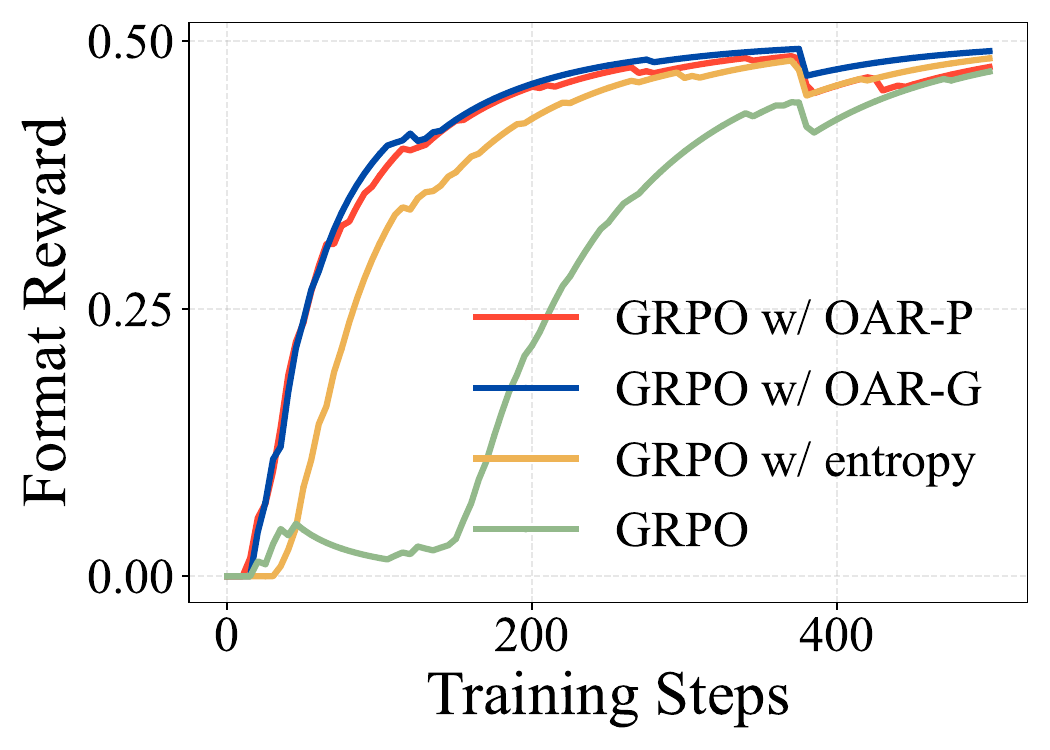}
\end{minipage}\hfill
\begin{minipage}{0.25\textwidth}\centering
  \includegraphics[width=\linewidth]{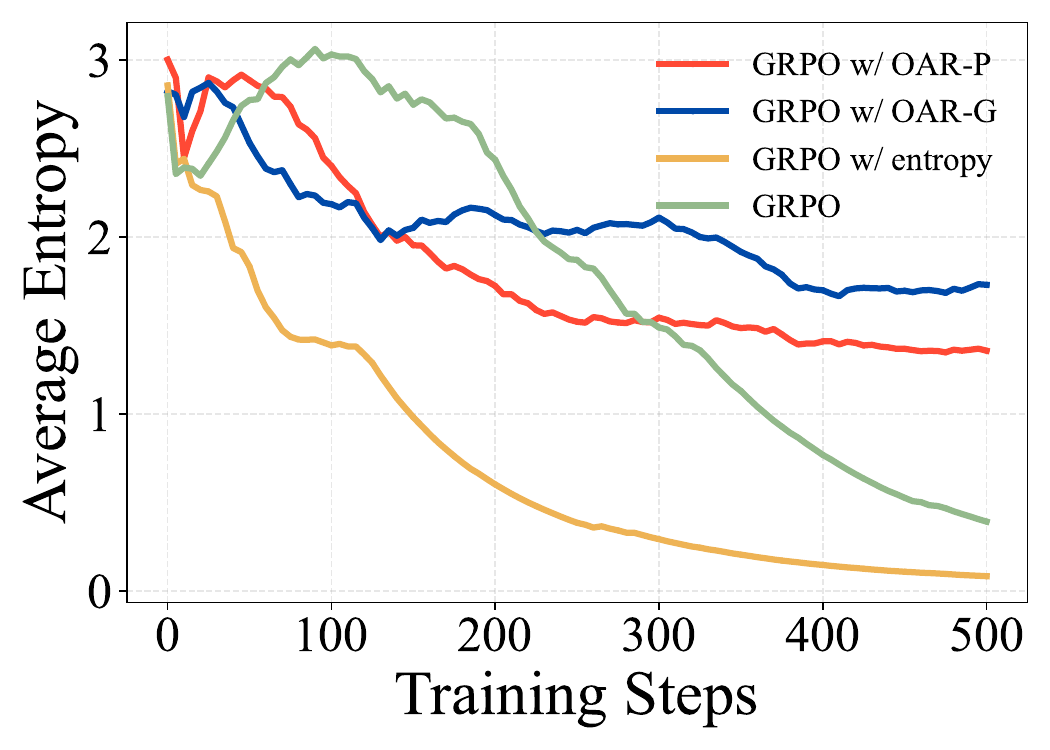}
\end{minipage}

\vspace{4pt}

\begin{minipage}{0.25\textwidth}\centering
  \includegraphics[width=\linewidth]{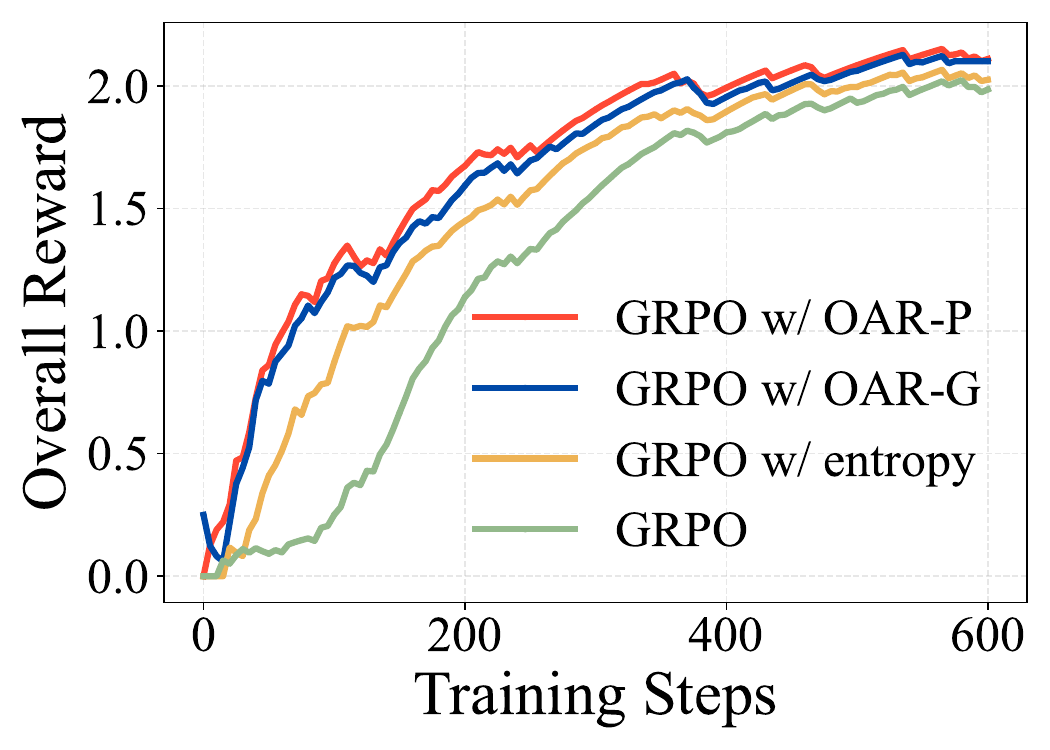}
\end{minipage}\hfill
\begin{minipage}{0.25\textwidth}\centering
  \includegraphics[width=\linewidth]{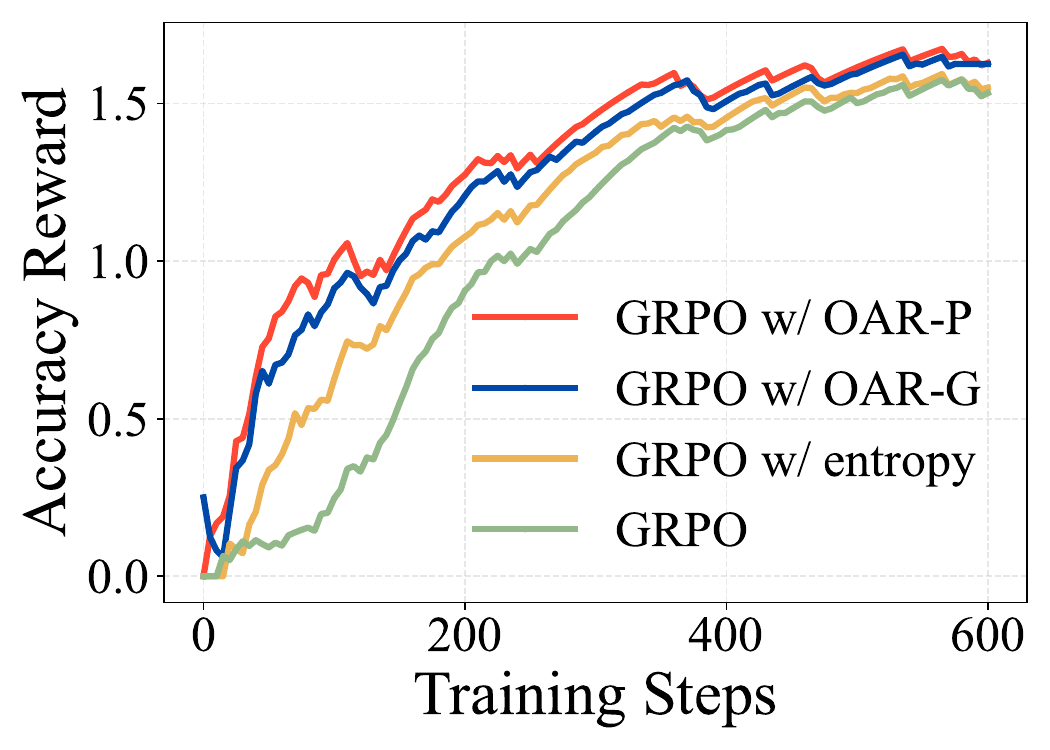}
\end{minipage}\hfill
\begin{minipage}{0.25\textwidth}\centering
  \includegraphics[width=\linewidth]{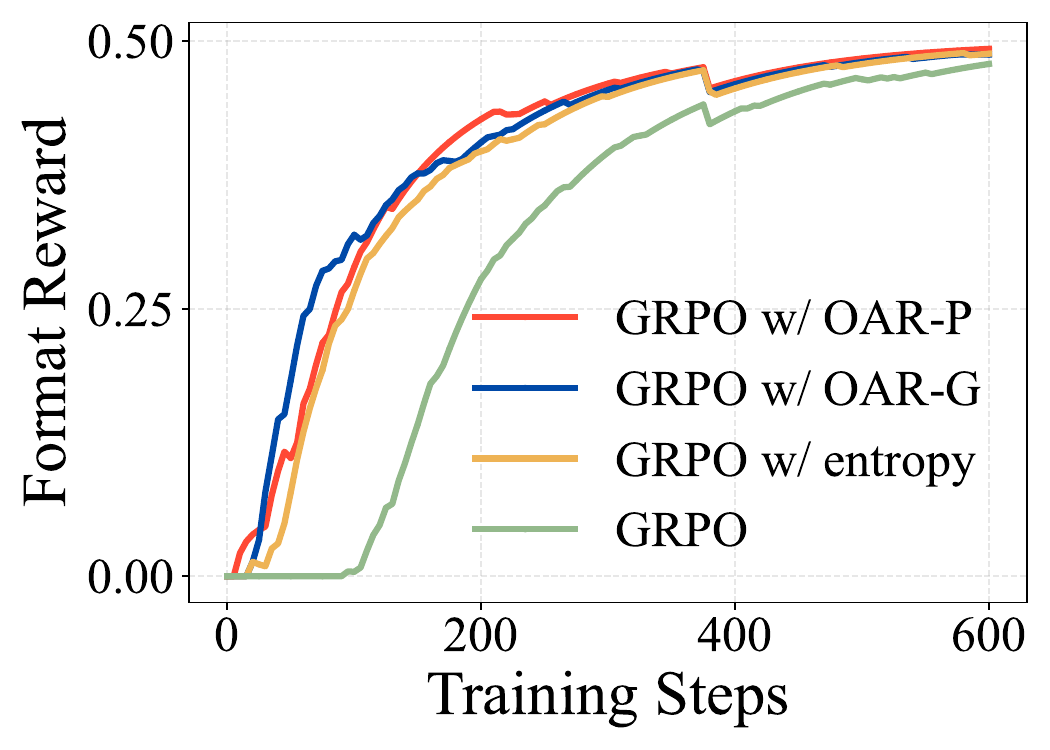}
\end{minipage}\hfill
\begin{minipage}{0.25\textwidth}\centering
  \includegraphics[width=\linewidth]{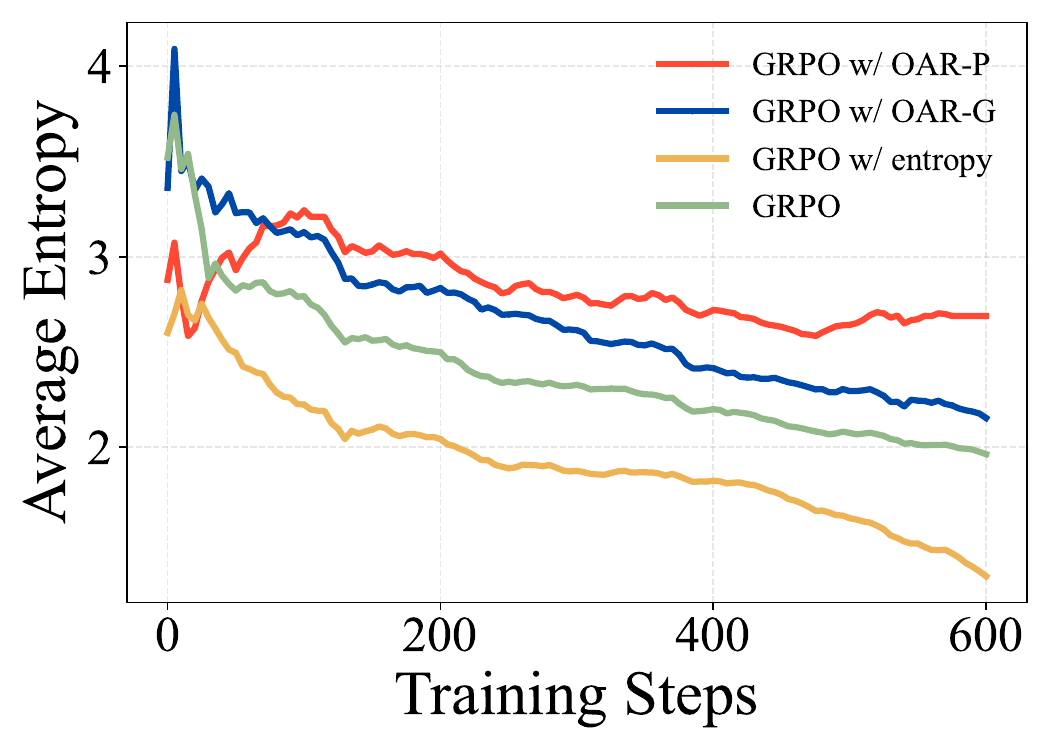}
\end{minipage}

\caption{Training dynamics on Qwen2.5-7B-Base(Top Row) and Qwen2.5-Math-7B(Bottom Row).}
\label{fig:training_dynamics}
\end{figure*}

\subsection{Bi-Level Advantage Reshaping}
\label{subsec:reshaping}

While the attribution scores $I_t$ identify pivotal tokens, directly applying them as raw weights can destabilize the training process due to uncontrolled variance in update scales.
To concentrate credit on important positions without changing GRPO's overall update scale, we reallocate $A_{\text{seq}}$ across tokens by suppressing low-importance tokens and boosting high-importance ones, followed by a sum-preserving renormalization. 
The token-level advantages are reshaped using the normalized importance $\hat{I}_t$. Specifically, we modulate the sequence advantage by a bi-level gating function:
\begin{equation}
\label{eq:par_advantage}
    A^{\text{OAR}}_t = A_{\text{seq}} \cdot \tilde{\omega}_t,
\end{equation}
where $\tilde{\omega}_t$ suppresses low-importance tokens and boosts high-importance tokens. Given a threshold $\tau$ (e.g., the 70th percentile in a sequence), we define
\begin{equation}
\label{eq:weighting_function}
    \omega(\hat{I}_t) =
    \begin{cases} 
        \underbrace{\frac{\hat{I}_t}{\tau + \epsilon}}_{\text{Noise Suppression}}, & \text{if } \hat{I}_t < \tau \\[15pt]
        \underbrace{1 + \beta \cdot \frac{\hat{I}_t - \tau}{1 - \tau + \epsilon}}_{\text{Signal Boosting}}, & \text{if } \hat{I}_t \ge \tau
    \end{cases}
\end{equation}
with boosting coefficient $\beta \ge 0$ and smoothing constant $\epsilon$. This function is continuous at $\hat{I}_t=\tau$ with $\omega(\tau)=1$. To avoid changing the overall update scale, we apply sum-preserving renormalization:
\begin{equation}
\label{eq:sum_preserving_renorm}
\tilde{\omega}_t \;=\; \omega(\hat{I}_t)\cdot
\frac{T}{\sum_{j=1}^{T}\omega(\hat{I}_j)}.
\end{equation}
This conserves total advantage mass ($\sum_t \tilde{\omega}_t = T$), so OAR redistributes credit across tokens rather than globally rescaling advantages.

Finally, we replace the standard advantage in the clipped PPO objective with $A^{\text{OAR}}_t$:
\begin{equation}
\begin{split}
    \mathcal{L}_{\text{OAR}} = \mathbb{E}_{i,t}\Bigg[ &
        \min\biggl( 
            \rho^{(i)}_t A^{\text{OAR}(i)}_t, \\
        & \quad
            \operatorname{clip}\bigl( \rho^{(i)}_t, 1-\epsilon, 1+\epsilon \bigr) A^{\text{OAR}(i)}_t 
        \biggr) 
    \Bigg].
\end{split}
\end{equation}

By integrating outcome-grounded credit into the surrogate loss, $\mathcal{L}_{\text{OAR}}$ encourages the policy to prioritize the refinement of critical reasoning steps while remaining robust to the high-variance noise inherent in long-horizon mathematical traces.

\begin{figure*}[t]
    \centering
    \includegraphics[width=\textwidth]{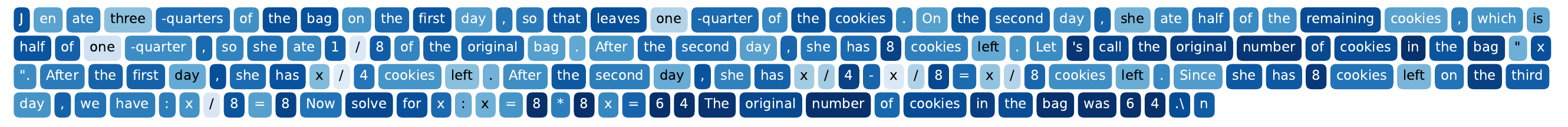}
    \includegraphics[width=\textwidth]{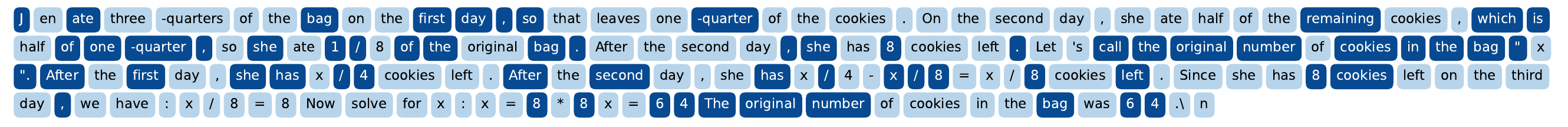}
    \caption{Token-importance visualization on a reasoning trace: OAR-P (top) vs.\ Oracle causal mask (bottom).}
    \label{fig:qualitative_map}
\end{figure*}

\section{Experiments}
\label{sec:experiments}

\subsection{Experiment Setup}
\label{subsec:setup}

\paragraph{Datasets and Models}
Experiments are conducted on \texttt{Qwen2.5-7B-Base} and \texttt{Qwen2.5-Math-7B}~\cite{yang2025qwen251mtechnicalreport}, trained with \texttt{DAPO-MATH} \cite{yu2025dapoopensourcellmreinforcement} dataset, We evaluate the resulting performance on four widely used math reasoning benchmarks:
AIME2024/2025 \cite{numina_math_datasets}, AMC23 \cite{numina_math_datasets}, MATH500 \cite{Hendrycks2021MeasuringMP}, and GSM8K \cite{Cobbe2021TrainingVT}.
For AIME and AMC, we report Pass@$k$ where $k \in \{1, 32\}$; for others we report standard Pass@1\citep{Chen2021EvaluatingLL}.

\paragraph{Implementation Details}
We employ the GRPO framework with a group size of $G=8$, learning rate of $1\times 10^{-6}$ and a global batch size of 64.
For generation, We set the temperature to 1.0 with a max generation length of 1024.
For our proposed OAR method, we set the bi-level threshold $\tau=0.4$ (meaning the top 60\% of tokens are boosted) and the boosting coefficient $\beta=2.0$.
To accelerate the counterfactual perturbation step during training, we implement a parallelized batch decoding strategy, which computes the perturbed logits for all masked positions in a single batched forward pass. All experiments are conducted on $8 \times \text{H100}$ GPUs.

\subsection{Baselines}
To validate the effectiveness of our method, we compare OAR against three baselines: (i) \emph{Vanilla GRPO}, the standard Group Relative Policy Optimization algorithm with a clip-higher technique from DAPO \cite{yu2025dapoopensourcellmreinforcement} as a strong reference; (ii) \emph{GRPO + Random Credit}, a diagnostic variant that assigns each token a random weight $w \sim \mathrm{Uniform}(0,1)$; (iii) \emph{GRPO + Entropy}, which follows \citet{cheng2025reasoningexplorationentropyperspective} and uses token entropy as an importance proxy to reshape advantages; and (iv) \emph{GRPO + KTAE}, which follows \citet{sun2025ktaemodelfreealgorithmkeytokens} and adds model-free token-level advantage corrections based on token--correctness association.

\subsection{Main Results}

\paragraph{Reasoning Performance}
Table~\ref{tab:main_results} summarizes results on five mathematical reasoning benchmarks using Qwen2.5-7B and Qwen2.5-Math-7B as backbones. Incorporating our OAR into GRPO consistently improves performance over vanilla GRPO and alternative token-weighting heuristics. For example, on Qwen2.5-7B, OAR-P improves AIME25 \textit{P@1} from 9.4 to 11.8 and boosts the overall average from 51.3 to \textbf{53.7}. Similar trends hold for Qwen2.5-Math-7B, where OAR-P raises the average from 57.3 to \textbf{59.5}.

\paragraph{Training Dynamics}
Figure~\ref{fig:training_dynamics} compares training dynamics under different credit assignment strategies. Across runs, OAR improves optimization stability and reaches higher rewards with fewer steps while avoiding premature entropy collapse. 

\begin{figure}[t]
    \centering
    \begin{subfigure}[t]{0.48\linewidth}
        \centering
        \includegraphics[width=\linewidth]{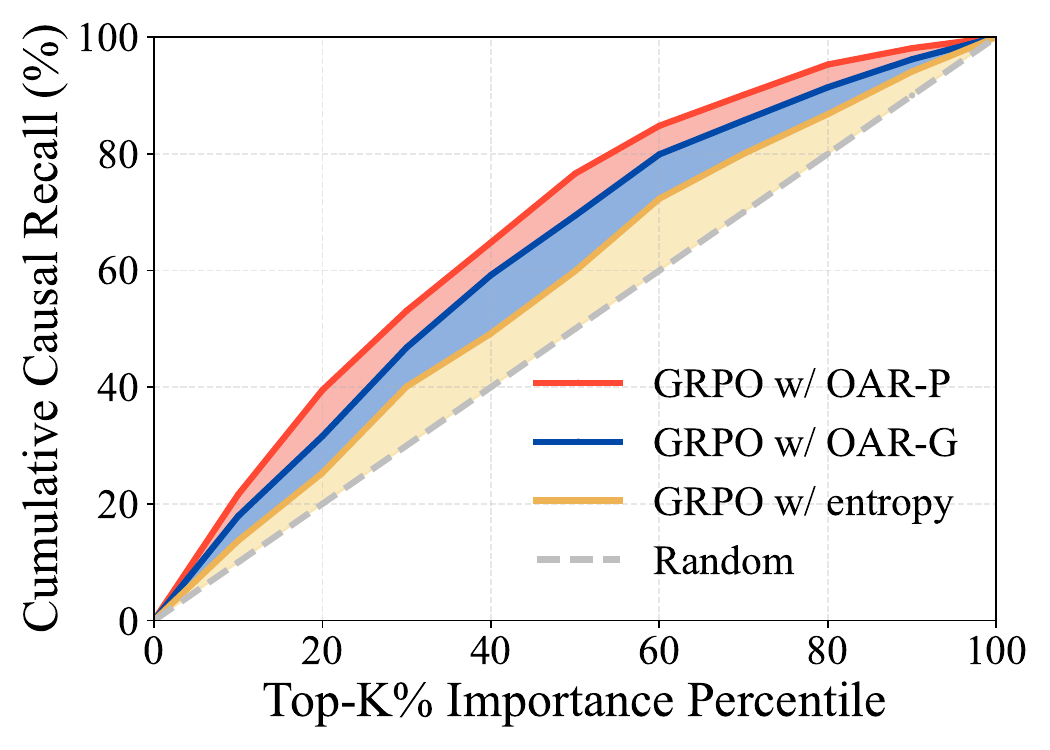}
        \caption{Recall rate of different credit assignment signals.}
        \label{fig:causal_recall:a}
    \end{subfigure}\hfill
    \begin{subfigure}[t]{0.48\linewidth}
        \centering
        \includegraphics[width=\linewidth]{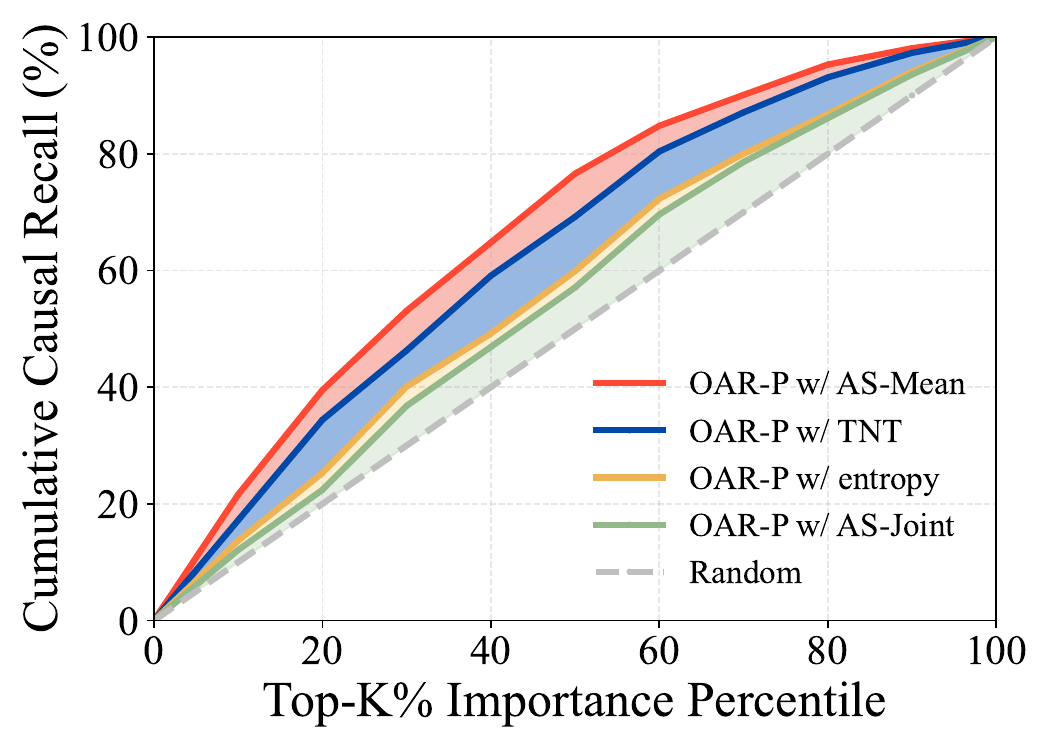}
        \caption{Recall rate under different outcome definitions.}
        \label{fig:causal_recall:b}
    \end{subfigure}
    \caption{Causal-token recall under the counterfactual Oracle as a function of the top-$K\%$ important tokens.}
    \label{fig:causal_recall}
\end{figure}

\begin{figure*}[t]
\centering
    \begin{minipage}{0.25\textwidth}\centering
      \includegraphics[width=\linewidth]{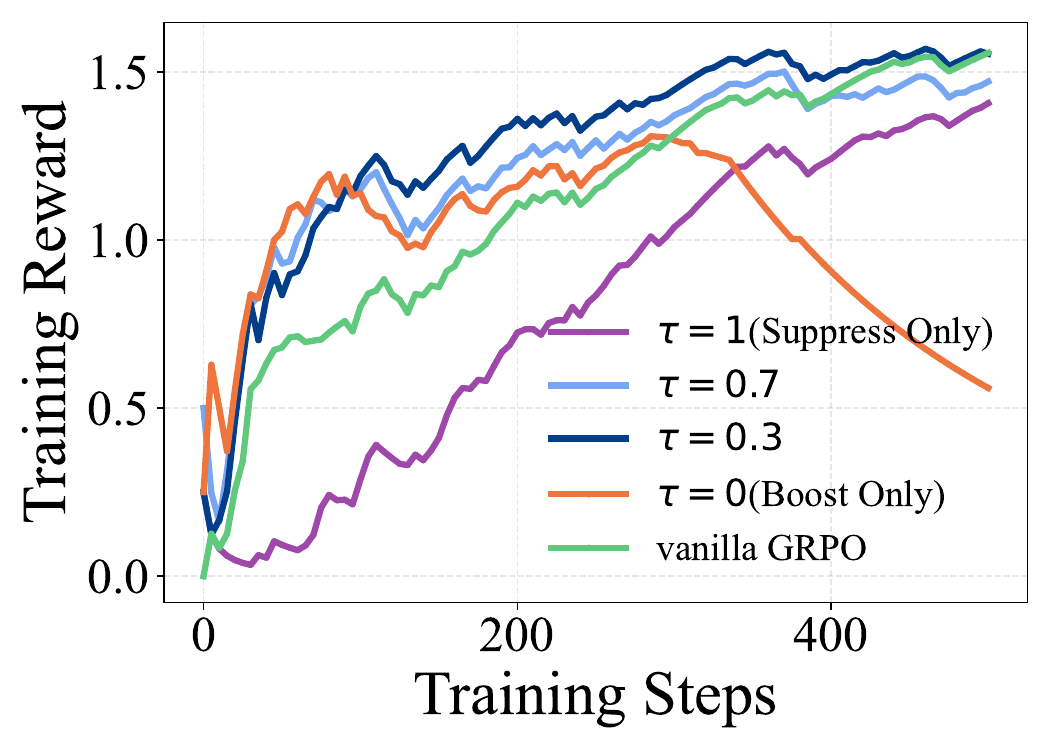}
    \end{minipage}\hfill
    \begin{minipage}{0.25\textwidth}\centering
      \includegraphics[width=\linewidth]{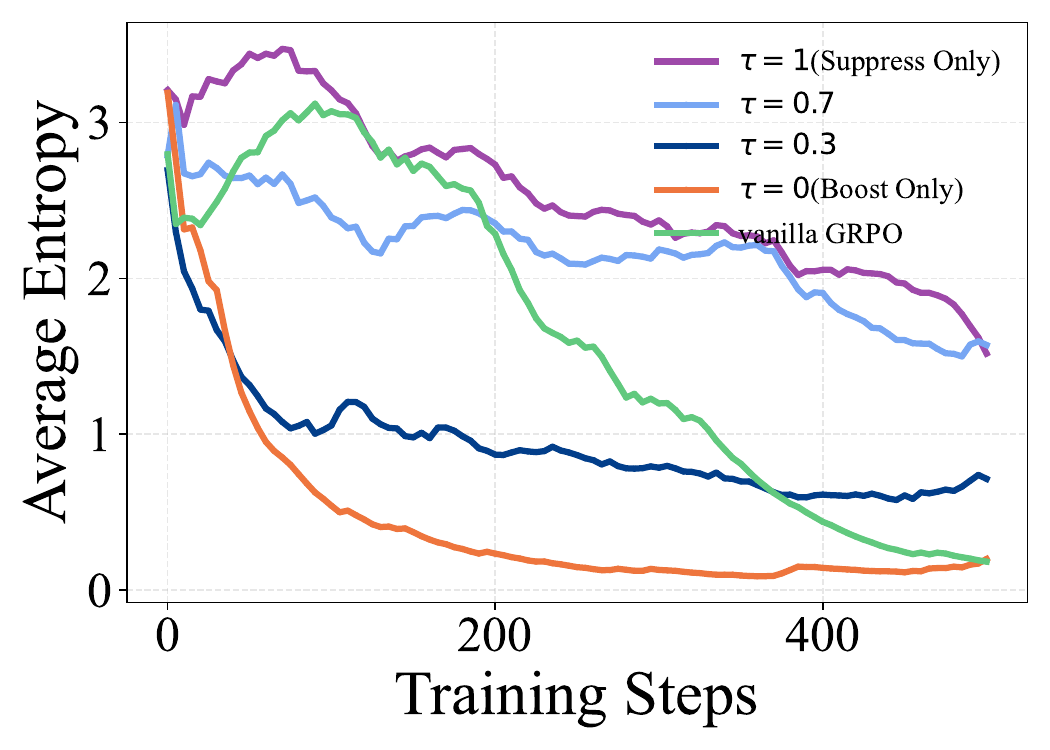}
    \end{minipage}\hfill
    \begin{minipage}{0.25\textwidth}\centering
      \includegraphics[width=\linewidth]{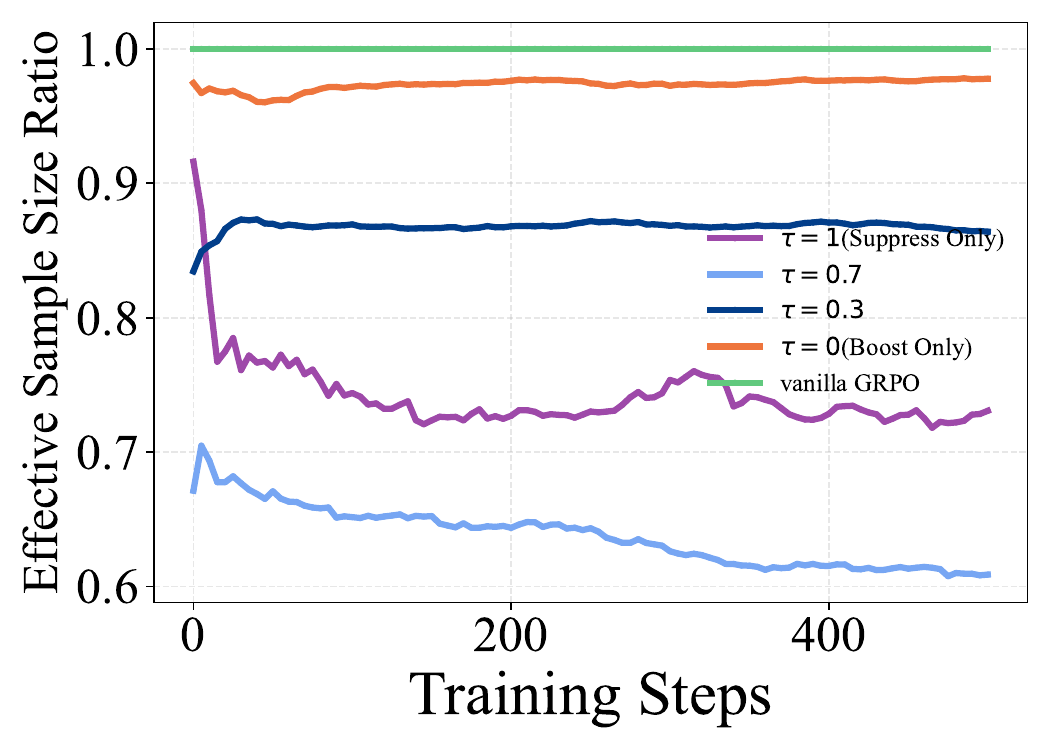}
    \end{minipage}\hfill
    \begin{minipage}{0.25\textwidth}\centering
      \includegraphics[width=\linewidth]{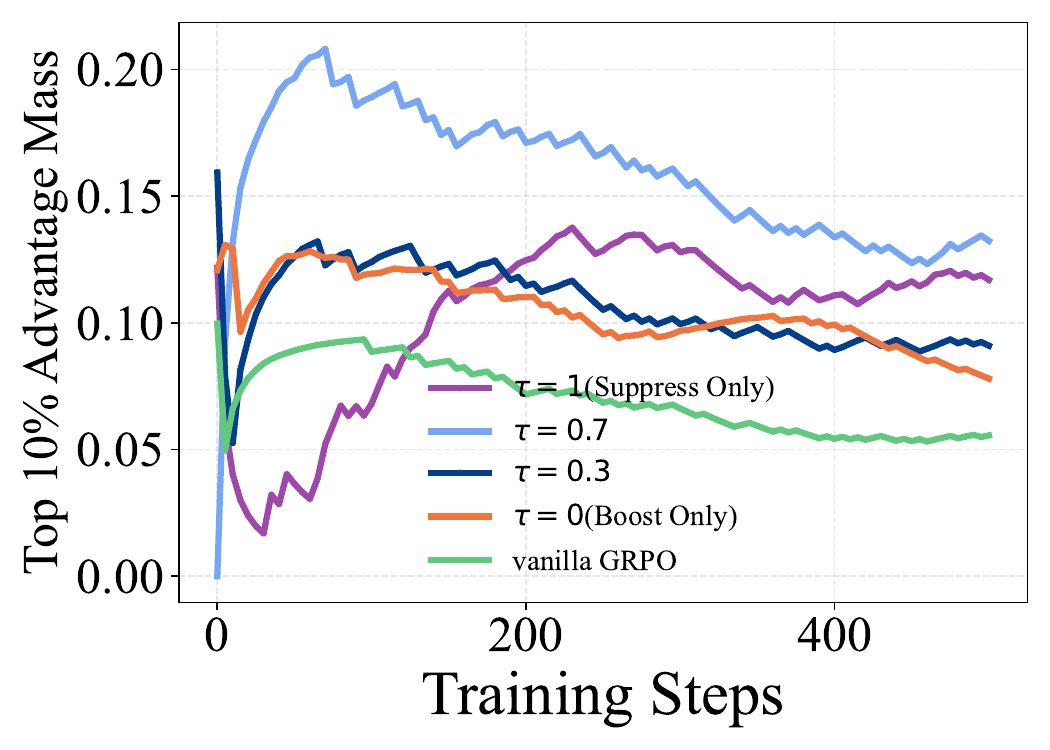}
    \end{minipage}
    \caption{OAR-G ablations on Qwen2.5-7B-Base with different thresholds $\tau$ (left to right): reward, policy entropy, ESS ratio, and top-10\% advantage mass.}
    \label{fig:ablation_dynamics}
\end{figure*}

\paragraph{Scaling to Different Sizes}
We additionally conduct experiments on a smaller \textit{Qwen2.5-Math-1.5B} model to verify the consistency and scalability of OAR. As shown in Table~\ref{tab:main_results_qwen25_math_1_5b}, OAR-G and OAR-P consistently improve over baselines.

\section{Analysis}
\subsection{Is OAR Identifying Important Tokens?}
\label{subsec:causal_validation}

\begin{table}[t]
\centering
\small
\setlength{\tabcolsep}{3pt}
\renewcommand{\arraystretch}{1}
\begin{tabular}{lcccc}
\toprule
Method & AIME24 & AMC23 & MATH500 & Avg \\
\midrule
\textit{Qwen2.5-Math-1.5B} & 2.9 & 20.2 & 40.3 & 21.1 \\
+ GRPO & 13.8 & 56.7 & 73.0 & 47.8 \\
w/ Random Adv & 13.8 & 57.0 & 72.6 & 47.8 \\
w/ Entropy Adv & 14.2 & \underline{58.2} & 74.5 & 48.9 \\
\rowcolor{blue!5}
w/ OAR-G & \underline{14.4} & 58.0 & \underline{75.1} & \underline{49.2} \\
\rowcolor{blue!10}
w/ OAR-P & \textbf{15.0} & \textbf{58.7} & \textbf{75.5} & \textbf{49.7} \\
\midrule
$\Delta$ (vs. GRPO) & \textcolor{green}{+1.2} & \textcolor{green}{+2.0} & \textcolor{green}{+2.5} & \textcolor{green}{+1.9} \\
\bottomrule
\end{tabular}
\caption{Results on Qwen2.5-Math-1.5B.}
\label{tab:main_results_qwen25_math_1_5b}
\end{table}

A core premise of OAR is that the identified token importance score $\hat{I}_t$ correlates with a token’s true causal contribution to the final outcome. We test this premise on Qwen2.5-7B-Base using a counterfactual Oracle and evaluate both OAR-P and OAR-G against entropy-based weighting in a quantitative correlation study.

\paragraph{The Oracle}
Given a correct reasoning chain $y=(y_1,\dots,y_T)$, following the counterfactual token-substitution test used to identify critical tokens in~\cite{ruan2025enhancinglargelanguagemodel},  we construct a counterfactual sequence by replacing token $y_t$ with the second most probable token under $\pi_\theta(\cdot\mid y_{<t})$ , and then greedily decoding the remaining suffix. Let $\mathcal{R}(\cdot)\in\{0,1\}$ denote final-answer correctness. The Oracle label is:
\begin{equation}
    \mathcal{O}_t = \mathbb{I}\left( \mathcal{R}(y_{\text{replaced}}^{(t)}) \neq \mathcal{R}(y_{\text{original}}) \right).
\end{equation}
If replacement flips correctness, $y_t$ is a \textbf{Causal Pivot} ($\mathcal{O}_t=1$); otherwise it is \textbf{Non-Causal Noise} ($\mathcal{O}_t=0$).

\paragraph{Alignment Visualization}
Figure~\ref{fig:qualitative_map} visualizes token importance for OAR-P (top) alongside the Oracle mask (bottom). OAR-P assigns high importance to outcome-critical symbols while down-weighting stylistic or redundant tokens, qualitatively matching the sparse counterfactual supervision. We provide additional case studies for OAR-G in Appendix ~\ref{sec:appendix:token_vis}.

\paragraph{Quantitative Correlation}
We sample 500 trajectories, rank tokens by $\hat{I}_t$, and compute the recall of Oracle causal tokens within the top-$K\%$ tokens. In Figure~\ref{fig:causal_recall}a, both OAR-P and OAR-G achieve substantially higher causal-token recall than entropy-based weighting across all percentiles, with OAR-P consistently performing best.

\begin{figure}
    \centering
    \includegraphics[width=0.98\linewidth]{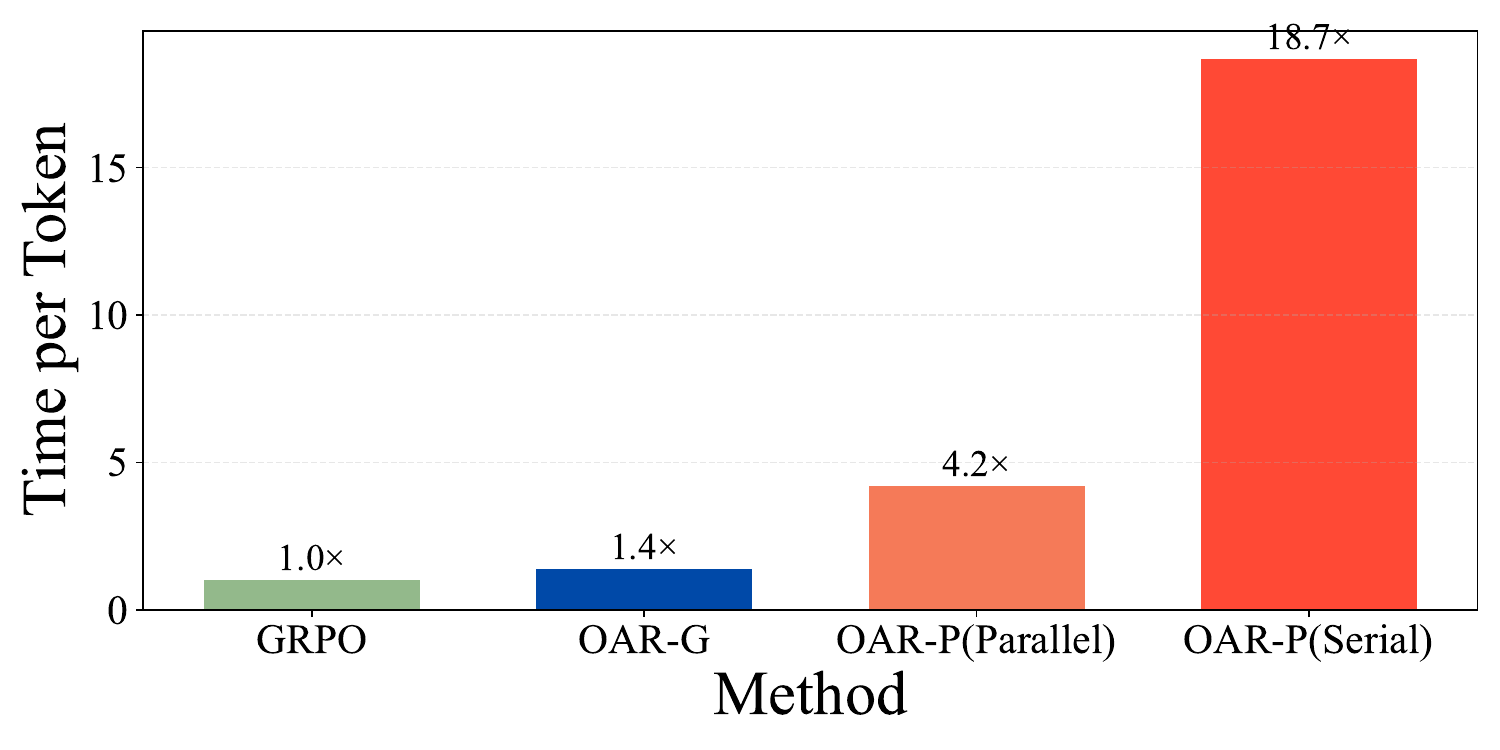}
    \caption{Normalized training time per token.}
    \label{fig:tradeoff_tpt}
\end{figure}

\subsection{Computational Trade-off}
\label{subsec:trade-off}
OAR improves credit assignment at the cost of additional attribution computation. We quantify this overhead using wall-clock time per token during training. For each update iteration, we record the elapsed time $\Delta t$ of one optimizer-update sweep over the buffered minibatches, and divide it by the number of action tokens.

Figure~\ref{fig:tradeoff_tpt} shows the normalized cost (GRPO = $1.0\times$). OAR-G incurs only a modest overhead ($1.4\times$), consistent with requiring a single backward-based proxy for importance. In contrast, OAR-P is substantially more expensive due to counterfactual evaluation: even with our batched (parallel) implementation it costs $4.2\times$, while a naive serial implementation would be prohibitive ($18.7\times$). Overall, OAR-G offers the best accuracy--efficiency balance, while OAR-P serves as a higher-fidelity but costlier upper bound.

\section{Ablations}
\subsection{Effect of the Bi-Level Gating Mechanism}
\label{subsec:ablation}
\begin{table}[h]
    \centering
    \small
    \begin{tabular}{l|cc}
    \toprule
    \textbf{Method} & \textbf{GSM8K} & \textbf{MATH500} \\
    \midrule
    Vanilla GRPO & 90.5 & 75.8 \\
    \midrule
    OAR (Suppress-only, $\tau=1.0$) & 85.2 & 70.1 \\
    OAR (Boost-only, $\tau=0.0$) & 81.5$^\dagger$ & 61.3$^\dagger$ \\
    OAR (Balanced, $\tau=0.7$) & 89.8 & 74.2 \\
    \textbf{OAR (Balanced, $\tau=0.3$)} & \textbf{91.2} & \textbf{78.1} \\
    \bottomrule
    \end{tabular}
    \caption{Ablation results on Qwen2.5-7B-Base. $^\dagger$ indicates runs that exhibited instability or collapse during training.}
    \label{tab:ablation_acc}
\end{table}

We ablate the bi-level gating in OAR by varying the threshold $\tau$ on Qwen2.5-7B-Base. This study is conducted based on OAR-G, for brevity we refer to it as OAR. Figure~\ref{fig:ablation_dynamics} reports training dynamics, and Table~\ref{tab:ablation_acc} summarizes final accuracy.

\paragraph{Credit concentration}
We evaluate whether OAR indeed reallocates token-level credit using two auxiliary metrics (full definitions in Appendix~\ref{app:credit_metrics}). \textit{ESS ratio} measures how uniform the token weights are (smaller means more concentrated), while \textit{Top-10\% advantage mass} measures what fraction of total token advantage lies in the top 10\% tokens. Compared with vanilla GRPO, OAR yields markedly more concentrated credit assignment, and larger $\tau$ generally produces sparser allocations.

\paragraph{Trade-off}
More aggressive sparsification is not always beneficial. As shown by the reward and entropy curves in Figure~\ref{fig:ablation_dynamics}, a moderate threshold performs best. \textit{Boost-only} ($\tau=0$) learns quickly but is prone to collapse, whereas \textit{suppress-only} ($\tau=1$) remains stable yet inefficient. Overall, the balanced setting achieves the best accuracy (Table~\ref{tab:ablation_acc}).

\subsection{Effect of Outcome Definition}
\label{subsec:ablation_outcome_def}

OAR is outcome-grounded through an outcome probe $\phi(x,y)$ that summarizes the model’s final-answer prediction. 
Because $\phi$ is not unique, different definitions may induce different token-importance rankings and thus different credit reallocation behaviors. 
We therefore ablate several outcome definitions to assess the robustness of causal-token identification and its impact on downstream RL performance.

\paragraph{Outcome definitions}
We consider three outcome operators $\phi(\cdot)$:
\begin{itemize}[itemsep=2pt, parsep=0pt, topsep=2pt]
\raggedright
    \item \textbf{Last-Token Logits (LT-Logits).}
    $\phi_{\textsc{lt-logits}}(x,y)=z_L$, i.e., the next-token logits at the last position.

    \item \textbf{Answer-Span Mean Logits (AS-Mean).}
    $\phi_{\textsc{as-mean}}(x,y)=\frac{1}{|\mathcal{A}|}\sum_{\ell\in\mathcal{A}} z_\ell$, averaging logits over positions $\mathcal{A}$ that predict the final answer span, where the span is extracted by regex matching \texttt{<answer>...</answer>} in the model output.

    \item \textbf{Answer-Span Joint Likelihood (AS-Joint).}
    $\phi_{\textsc{as-joint}}(x,y)=\frac{1}{|\mathcal{A}|}\sum_{\ell\in\mathcal{A}}\log\pi_\theta(a_{\ell+1}\!\mid x,y_{\le \ell})$, the mean token log-likelihood of the extracted answer span.
\end{itemize}

For LT-Logits and AS-Mean we compute importance via $D_{\mathrm{KL}}(\mathrm{softmax}(\phi)\,\|\,\mathrm{softmax}(\tilde\phi))$ under token masking; for AS-Joint we use the score drop $\phi-\tilde\phi$.

\paragraph{Experiment setup}
We follow the quantitative correlation setup in Section~\ref{subsec:causal_validation}, measure the recall rate under three methods. Notability, For \textsc{AS-Mean}, both in this experiment and in our training runs, we initially apply \textsc{LT-Logits} as a warm-up, as in early training stage the model may not reliably produce the required \textit{<answer>...</answer>} format. We further report the task accuracy to connect attribution quality with actual performance.

\begin{table}[t]
    \centering
    \small
    \begin{tabular}{l|cc}
    \toprule
    \textbf{Outcome definition} & \textbf{GSM8K} & \textbf{MATH500} \\
    \midrule
    \textsc{LT-Logits}      & 91.5 & 78.0 \\
    \textsc{AS-Joint} & 91.0 & 77.3 \\
    \textsc{AS-Mean}  & \textbf{92.0} & \textbf{78.4} \\
    \bottomrule
    \end{tabular}
    \caption{Accuracy ablation of outcome definitions.}
    \label{tab:ablation_outcome_acc}
\end{table}

\paragraph{Results}
Figure~\ref{fig:causal_recall}b and Table~\ref{tab:ablation_outcome_acc} show that \textsc{AS-Mean} achieves the highest Oracle top-$K\%$ recall and the best downstream training performance, followed by \textsc{LT-Logits}, while \textsc{AS-Joint} performs worst. We attribute this to a representational mismatch: \textsc{LT-Logits} and \textsc{AS-Mean} operate on distributions and thus can capture subtle shifts in the outcome distribution induced by token perturbations, whereas \textsc{AS-Joint} collapses the outcome into a scalar likelihood and may miss substantial distributional changes.

\section{Related Work}
\paragraph{Reinforcement Learning with Verifiable Rewards}
RLVR post-trains LLMs using automatically checkable outcome signals, yielding strong gains in math and code via verifiers such as Math-Verify and sandboxed execution \citep{cui2025processreinforcementimplicitrewards,yu2025dapoopensourcellmreinforcement,deepscaler2025,deepseekai2025deepseekr1incentivizingreasoningcapability}. 
GRPO further stabilize large-scale training by group-normalizing verifier rewards \citep{Shao2024DeepSeekMathPT}. 
To move beyond domains with explicit checkers, recent studies learn generative reward/verifier models for model-based judging \citep{mahan2024generativerewardmodels,ma2025generalreasoneradvancingllmreasoning,liu2025understandingr1zeroliketrainingcritical}, while verifier-free variants instead use intrinsic signals such as per-token likelihood on reference answers to enable RL-style post-training without dedicated verifiers \citep{yu2025rlprextrapolatingrlvrgeneral,zhou2025reinforcinggeneralreasoningverifiers}.

\paragraph{Fine-Grained Credit Assignment}
To address the coarseness of group-level rewards, recent studies explored leveraging model-internal signals to redistribute credits at the token level.
\citet{li2025attention} identify a recurring \emph{preplan-and-anchor} rhythm from attention dynamics and upweight advantages on the corresponding critical tokens.
\citet{xie2025unlockingexplorationrlvruncertaintyaware} propose UCAS, which reshapes advantages using response-level confidence and a token-level logit-based certainty penalty to encourage exploration.
\citet{chen2025highentropyexplorationcorrectnessawarelowentropy} introduce LESS, which segments generations by entropy and reweights advantages based on low-entropy segment overlap between correct and incorrect rollouts.


\section{Conclusion}
We propose Outcome-grounded Advantage Reshaping (OAR) to improve credit assignment in GRPO for long-horizon reasoning. OAR redistributes sequence-level advantages to tokens based on outcome contribution, with OAR-P using counterfactual attribution and OAR-G a lightweight gradient-based proxy. Experiments on mathematical reasoning benchmarks show consistent gains over strong GRPO baselines. Future work will extend OAR to open-ended tasks and deeper RL post-training integration.

\section*{Limitaions}
OAR estimates token importance by measuring shifts in the model's own answer distribution, which serves as an efficient surrogate when verifier rewards are discrete and non-differentiable; nevertheless, this proxy can be imperfect when distributional changes are weakly coupled with verifier acceptance. Our perturbation-based attribution further relies on a masking intervention (e.g., \texttt{[PAD]}), which is a controlled counterfactual but may introduce some distribution shift compared to in-distribution edits. Empirically, we focus on mathematical reasoning with verifiable final answers, and extending OAR to open-ended or interactive settings where outcomes are less localized remains an important direction. Finally, beyond entropy-based shaping, several closely related recent methods are not yet fully reproducible, we will broaden comparisons as these baselines become easier to faithfully re-implement.

\section*{Acknowledgements}
The authors would like to thank Fangyuan Mao for his assistance with the professional illustrations and diagrams in this paper. We also thank Shun Lu for his insightful reviews and valuable suggestions that significantly improved the quality of this work.

\bibliography{main}

@misc{Shao2024DeepSeekMathPT,
  author        = {Shao, Zhihong and Wang, Peiyi and Zhu, Qihao and Xu, Rui and Song, Junmei and Zhang, Mingchuan and Li, Y.K. and Wu, Yuxiang and Guo, Daya},
  title         = {{DeepSeekMath: Pushing the Limits of Mathematical Reasoning in Open Language Models}},
  year          = {2024},
  eprint        = {2402.03300},
  archiveprefix = {arXiv},
  primaryclass  = {cs.LG}
}

@misc{lambert2024tulu,
  author        = {Lambert, Nathan and Morrison, Jacob and Pyatkin, Valentina and Huang, Shengyi and Ivison, Hamish and Brahman, Faeze and Miranda, Lester James V. and Liu, Alisa and Dziri, Nouha and Lyu, Shane and Havasi, David and Prabhumoye, Shrimai and Liu, Pengfei and Dodge, Jesse and Choi, Yejin and Smith, Noah A. and Hajishirzi, Hannaneh and Zettlemoyer, Luke},
  title         = {{Tulu 3: Pushing frontiers in open language model post-training}},
  year          = {2024},
  eprint        = {2411.15124},
  archiveprefix = {arXiv},
  primaryclass  = {cs.CL}
}

@article{liu2024tis,
  title={Tis-dpo: Token-level importance sampling for direct preference optimization with estimated weights},
  author={Liu, Aiwei and Bai, Haoping and Lu, Zhiyun and Sun, Yanchao and Kong, Xiang and Wang, Simon and Shan, Jiulong and Jose, Albin Madappally and Liu, Xiaojiang and Wen, Lijie and others},
  journal={arXiv preprint arXiv:2410.04350},
  year={2024}
}

@article{li2024preserving,
  title={Preserving diversity in supervised fine-tuning of large language models},
  author={Li, Ziniu and Chen, Congliang and Xu, Tian and Qin, Zeyu and Xiao, Jiancong and Luo, Zhi-Quan and Sun, Ruoyu},
  journal={arXiv preprint arXiv:2408.16673},
  year={2024}
}

@article{achiam2023gpt4,
  title={{GPT-4 Technical Report}},
  author={Achiam, Josh and Adler, Steven and Agarwal, Sandhini and Ahmad, Lama and Akkaya, Ilge and Aleman, Florencia Leoni and Almeida, Diogo and Altenschmidt, Janko and Altman, Sam and An, Shayne and others},
  journal={arXiv preprint arXiv:2303.08774},
  year={2023}
}

@misc{li2025attention,
  author        = {Li, Yang and Dong, Zhichen and Sun, Yuhan and Wang, Weixun and Xiong, Shaopan and Luo, Yijia and Liu, Jiashun and Lu, Han and Wang, Jiamang and Su, Wenbo and Zheng, Bo and Yan, Junchi},
  title         = {Attention Illuminates {LLM} Reasoning: The Preplan-and-Anchor Rhythm Enables Fine-Grained Policy Optimization},
  year          = {2025},
  eprint        = {2510.13554},
  archiveprefix = {arXiv},
  primaryclass  = {cs.CL}
}

@misc{cheng2025reasoningexplorationentropyperspective,
      title={Reasoning with Exploration: An Entropy Perspective}, 
      author={Daixuan Cheng and Shaohan Huang and Xuekai Zhu and Bo Dai and Wayne Xin Zhao and Zhenliang Zhang and Furu Wei},
      year={2025},
      eprint={2506.14758},
      archivePrefix={arXiv},
      primaryClass={cs.CL},
      url={https://arxiv.org/abs/2506.14758}, 
}

@misc{yang2025qwen251mtechnicalreport,
      title={Qwen2.5-1M Technical Report}, 
      author={An Yang and Bowen Yu and Chengyuan Li and Dayiheng Liu and Fei Huang and Haoyan Huang and Jiandong Jiang and Jianhong Tu and Jianwei Zhang and Jingren Zhou and Junyang Lin and Kai Dang and Kexin Yang and Le Yu and Mei Li and Minmin Sun and Qin Zhu and Rui Men and Tao He and Weijia Xu and Wenbiao Yin and Wenyuan Yu and Xiafei Qiu and Xingzhang Ren and Xinlong Yang and Yong Li and Zhiying Xu and Zipeng Zhang},
      year={2025},
      eprint={2501.15383},
      archivePrefix={arXiv},
      primaryClass={cs.CL},
      url={https://arxiv.org/abs/2501.15383}, 
}

@misc{yu2025dapoopensourcellmreinforcement,
      title={DAPO: An Open-Source LLM Reinforcement Learning System at Scale}, 
      author={Qiying Yu and Zheng Zhang and Ruofei Zhu and Yufeng Yuan and Xiaochen Zuo and Yu Yue and Weinan Dai and Tiantian Fan and Gaohong Liu and Lingjun Liu and Xin Liu and Haibin Lin and Zhiqi Lin and Bole Ma and Guangming Sheng and Yuxuan Tong and Chi Zhang and Mofan Zhang and Wang Zhang and Hang Zhu and Jinhua Zhu and Jiaze Chen and Jiangjie Chen and Chengyi Wang and Hongli Yu and Yuxuan Song and Xiangpeng Wei and Hao Zhou and Jingjing Liu and Wei-Ying Ma and Ya-Qin Zhang and Lin Yan and Mu Qiao and Yonghui Wu and Mingxuan Wang},
      year={2025},
      eprint={2503.14476},
      archivePrefix={arXiv},
      primaryClass={cs.LG},
      url={https://arxiv.org/abs/2503.14476}, 
}

@Article{Chen2021EvaluatingLL,
 author = {Mark Chen and Jerry Tworek and Heewoo Jun and Qiming Yuan and Henrique Pondé and Jared Kaplan and Harrison Edwards and Yura Burda and Nicholas Joseph and Greg Brockman and Alex Ray and Raul Puri and Gretchen Krueger and Michael Petrov and Heidy Khlaaf and Girish Sastry and Pamela Mishkin and Brooke Chan and Scott Gray and Nick Ryder and Mikhail Pavlov and Alethea Power and Lukasz Kaiser and Mohammad Bavarian and Clemens Winter and Philippe Tillet and F. Such and D. Cummings and Matthias Plappert and Fotios Chantzis and Elizabeth Barnes and Ariel Herbert-Voss and William H. Guss and Alex Nichol and Igor Babuschkin and S. Balaji and Shantanu Jain and A. Carr and J. Leike and Joshua Achiam and Vedant Misra and Evan Morikawa and Alec Radford and M. Knight and Miles Brundage and Mira Murati and Katie Mayer and P. Welinder and Bob McGrew and Dario Amodei and Sam McCandlish and I. Sutskever and Wojciech Zaremba},
 booktitle = {arXiv.org},
 journal = {ArXiv},
 title = {Evaluating Large Language Models Trained on Code},
 volume = {abs/2107.03374},
 year = {2021}
}

@Article{Hendrycks2021MeasuringMP,
 author = {Dan Hendrycks and Collin Burns and Saurav Kadavath and Akul Arora and Steven Basart and Eric Tang and D. Song and J. Steinhardt},
 booktitle = {NeurIPS Datasets and Benchmarks},
 journal = {ArXiv},
 title = {Measuring Mathematical Problem Solving With the MATH Dataset},
 volume = {abs/2103.03874},
 year = {2021}
}

@Article{Cobbe2021TrainingVT,
 author = {K. Cobbe and Vineet Kosaraju and Mohammad Bavarian and Mark Chen and Heewoo Jun and Lukasz Kaiser and Matthias Plappert and Jerry Tworek and Jacob Hilton and Reiichiro Nakano and Christopher Hesse and John Schulman},
 booktitle = {arXiv.org},
 journal = {ArXiv},
 title = {Training Verifiers to Solve Math Word Problems},
 volume = {abs/2110.14168},
 year = {2021}
}

@misc{sundararajan2017axiomaticattributiondeepnetworks,
      title={Axiomatic Attribution for Deep Networks}, 
      author={Mukund Sundararajan and Ankur Taly and Qiqi Yan},
      year={2017},
      eprint={1703.01365},
      archivePrefix={arXiv},
      primaryClass={cs.LG},
      url={https://arxiv.org/abs/1703.01365}, 
}

@misc{ruan2025enhancinglargelanguagemodel,
      title={Enhancing Large Language Model Reasoning via Selective Critical Token Fine-Tuning}, 
      author={Zhiwen Ruan and Yixia Li and He Zhu and Yun Chen and Peng Li and Yang Liu and Guanhua Chen},
      year={2025},
      eprint={2510.10974},
      archivePrefix={arXiv},
      primaryClass={cs.CL},
      url={https://arxiv.org/abs/2510.10974}, 
}

@misc{numina_math_datasets,
  author = {Jia LI and Edward Beeching and Lewis Tunstall and Ben Lipkin and Roman Soletskyi and Shengyi Costa Huang and Kashif Rasul and Longhui Yu and Albert Jiang and Ziju Shen and Zihan Qin and Bin Dong and Li Zhou and Yann Fleureau and Guillaume Lample and Stanislas Polu},
  title = {NuminaMath},
  year = {2024},
  publisher = {Numina},
  journal = {Hugging Face repository},
  howpublished = {\url{[https://huggingface.co/AI-MO/NuminaMath-CoT](https://github.com/project-numina/aimo-progress-prize/blob/main/report/numina_dataset.pdf)}}
}

@misc{chen2025highentropyexplorationcorrectnessawarelowentropy,
      title={Beyond High-Entropy Exploration: Correctness-Aware Low-Entropy Segment-Based Advantage Shaping for Reasoning LLMs}, 
      author={Xinzhu Chen and Xuesheng Li and Zhongxiang Sun and Weijie Yu},
      year={2025},
      eprint={2512.00908},
      archivePrefix={arXiv},
      primaryClass={cs.LG},
      url={https://arxiv.org/abs/2512.00908}, 
}

@misc{fedus2022switchtransformersscalingtrillion,
      title={Switch Transformers: Scaling to Trillion Parameter Models with Simple and Efficient Sparsity}, 
      author={William Fedus and Barret Zoph and Noam Shazeer},
      year={2022},
      eprint={2101.03961},
      archivePrefix={arXiv},
      primaryClass={cs.LG},
      url={https://arxiv.org/abs/2101.03961}, 
}

@misc{brown2020languagemodelsfewshotlearners,
      title={Language Models are Few-Shot Learners}, 
      author={Tom B. Brown and Benjamin Mann and Nick Ryder and Melanie Subbiah and Jared Kaplan and Prafulla Dhariwal and Arvind Neelakantan and Pranav Shyam and Girish Sastry and Amanda Askell and Sandhini Agarwal and Ariel Herbert-Voss and Gretchen Krueger and Tom Henighan and Rewon Child and Aditya Ramesh and Daniel M. Ziegler and Jeffrey Wu and Clemens Winter and Christopher Hesse and Mark Chen and Eric Sigler and Mateusz Litwin and Scott Gray and Benjamin Chess and Jack Clark and Christopher Berner and Sam McCandlish and Alec Radford and Ilya Sutskever and Dario Amodei},
      year={2020},
      eprint={2005.14165},
      archivePrefix={arXiv},
      primaryClass={cs.CL},
      url={https://arxiv.org/abs/2005.14165}, 
}

@misc{xie2025unlockingexplorationrlvruncertaintyaware,
      title={Unlocking Exploration in RLVR: Uncertainty-aware Advantage Shaping for Deeper Reasoning}, 
      author={Can Xie and Ruotong Pan and Xiangyu Wu and Yunfei Zhang and Jiayi Fu and Tingting Gao and Guorui Zhou},
      year={2025},
      eprint={2510.10649},
      archivePrefix={arXiv},
      primaryClass={cs.AI},
      url={https://arxiv.org/abs/2510.10649}, 
}

@misc{cui2025processreinforcementimplicitrewards,
      title={Process Reinforcement through Implicit Rewards}, 
      author={Ganqu Cui and Lifan Yuan and Zefan Wang and Hanbin Wang and Yuchen Zhang and Jiacheng Chen and Wendi Li and Bingxiang He and Yuchen Fan and Tianyu Yu and Qixin Xu and Weize Chen and Jiarui Yuan and Huayu Chen and Kaiyan Zhang and Xingtai Lv and Shuo Wang and Yuan Yao and Xu Han and Hao Peng and Yu Cheng and Zhiyuan Liu and Maosong Sun and Bowen Zhou and Ning Ding},
      year={2025},
      eprint={2502.01456},
      archivePrefix={arXiv},
      primaryClass={cs.LG},
      url={https://arxiv.org/abs/2502.01456}, 
}

@misc{deepscaler2025,
  title={{DeepScaleR: Surpassing O1-Preview with a 1.5B Model by Scaling RL}},
  author={Luo, Michael and Tan, Sijun and Wong, Justin and Shi, Xiaoxiang and Tang, William Y. and Roongta, Manan and Cai, Colin and Luo, Jeffrey and Li, Li Erran and Popa, Raluca Ada and Stoica, Ion},
  howpublished={\url{https://pretty-radio-b75.notion.site/DeepScaleR-Surpassing-O1-Preview-with-a-1-5B-Model-by-Scaling-RL-19681902c1468005bed8ca303013a4e2}},
  year={2025},
  note={Notion Blog}
}

@misc{deepseekai2025deepseekr1incentivizingreasoningcapability,
      title={DeepSeek-R1: Incentivizing Reasoning Capability in LLMs via Reinforcement Learning}, 
      author={DeepSeek-AI and Daya Guo and Dejian Yang and Haowei Zhang and Junxiao Song and Ruoyu Zhang and Runxin Xu and Qihao Zhu and Shirong Ma and Peiyi Wang and Xiao Bi and Xiaokang Zhang and Xingkai Yu and Yu Wu and Z. F. Wu and Zhibin Gou and Zhihong Shao and Zhuoshu Li and Ziyi Gao and Aixin Liu and Bing Xue and Bingxuan Wang and Bochao Wu and Bei Feng and Chengda Lu and Chenggang Zhao and Chengqi Deng and Chenyu Zhang and Chong Ruan and Damai Dai and Deli Chen and Dongjie Ji and Erhang Li and Fangyun Lin and Fucong Dai and Fuli Luo and Guangbo Hao and Guanting Chen and Guowei Li and H. Zhang and Han Bao and Hanwei Xu and Haocheng Wang and Honghui Ding and Huajian Xin and Huazuo Gao and Hui Qu and Hui Li and Jianzhong Guo and Jiashi Li and Jiawei Wang and Jingchang Chen and Jingyang Yuan and Junjie Qiu and Junlong Li and J. L. Cai and Jiaqi Ni and Jian Liang and Jin Chen and Kai Dong and Kai Hu and Kaige Gao and Kang Guan and Kexin Huang and Kuai Yu and Lean Wang and Lecong Zhang and Liang Zhao and Litong Wang and Liyue Zhang and Lei Xu and Leyi Xia and Mingchuan Zhang and Minghua Zhang and Minghui Tang and Meng Li and Miaojun Wang and Mingming Li and Ning Tian and Panpan Huang and Peng Zhang and Qiancheng Wang and Qinyu Chen and Qiushi Du and Ruiqi Ge and Ruisong Zhang and Ruizhe Pan and Runji Wang and R. J. Chen and R. L. Jin and Ruyi Chen and Shanghao Lu and Shangyan Zhou and Shanhuang Chen and Shengfeng Ye and Shiyu Wang and Shuiping Yu and Shunfeng Zhou and Shuting Pan and S. S. Li and Shuang Zhou and Shaoqing Wu and Shengfeng Ye and Tao Yun and Tian Pei and Tianyu Sun and T. Wang and Wangding Zeng and Wanjia Zhao and Wen Liu and Wenfeng Liang and Wenjun Gao and Wenqin Yu and Wentao Zhang and W. L. Xiao and Wei An and Xiaodong Liu and Xiaohan Wang and Xiaokang Chen and Xiaotao Nie and Xin Cheng and Xin Liu and Xin Xie and Xingchao Liu and Xinyu Yang and Xinyuan Li and Xuecheng Su and Xuheng Lin and X. Q. Li and Xiangyue Jin and Xiaojin Shen and Xiaosha Chen and Xiaowen Sun and Xiaoxiang Wang and Xinnan Song and Xinyi Zhou and Xianzu Wang and Xinxia Shan and Y. K. Li and Y. Q. Wang and Y. X. Wei and Yang Zhang and Yanhong Xu and Yao Li and Yao Zhao and Yaofeng Sun and Yaohui Wang and Yi Yu and Yichao Zhang and Yifan Shi and Yiliang Xiong and Ying He and Yishi Piao and Yisong Wang and Yixuan Tan and Yiyang Ma and Yiyuan Liu and Yongqiang Guo and Yuan Ou and Yuduan Wang and Yue Gong and Yuheng Zou and Yujia He and Yunfan Xiong and Yuxiang Luo and Yuxiang You and Yuxuan Liu and Yuyang Zhou and Y. X. Zhu and Yanhong Xu and Yanping Huang and Yaohui Li and Yi Zheng and Yuchen Zhu and Yunxian Ma and Ying Tang and Yukun Zha and Yuting Yan and Z. Z. Ren and Zehui Ren and Zhangli Sha and Zhe Fu and Zhean Xu and Zhenda Xie and Zhengyan Zhang and Zhewen Hao and Zhicheng Ma and Zhigang Yan and Zhiyu Wu and Zihui Gu and Zijia Zhu and Zijun Liu and Zilin Li and Ziwei Xie and Ziyang Song and Zizheng Pan and Zhen Huang and Zhipeng Xu and Zhongyu Zhang and Zhen Zhang},
      year={2025},
      eprint={2501.12948},
      archivePrefix={arXiv},
      primaryClass={cs.CL},
      url={https://arxiv.org/abs/2501.12948}, 
}

@misc{mahan2024generativerewardmodels,
      title={Generative Reward Models}, 
      author={Dakota Mahan and Duy Van Phung and Rafael Rafailov and Chase Blagden and Nathan Lile and Louis Castricato and Jan-Philipp Fränken and Chelsea Finn and Alon Albalak},
      year={2024},
      eprint={2410.12832},
      archivePrefix={arXiv},
      primaryClass={cs.LG},
      url={https://arxiv.org/abs/2410.12832}, 
}

@misc{ma2025generalreasoneradvancingllmreasoning,
      title={General-Reasoner: Advancing LLM Reasoning Across All Domains}, 
      author={Xueguang Ma and Qian Liu and Dongfu Jiang and Ge Zhang and Zejun Ma and Wenhu Chen},
      year={2025},
      eprint={2505.14652},
      archivePrefix={arXiv},
      primaryClass={cs.CL},
      url={https://arxiv.org/abs/2505.14652}, 
}

@misc{liu2025understandingr1zeroliketrainingcritical,
      title={Understanding R1-Zero-Like Training: A Critical Perspective}, 
      author={Zichen Liu and Changyu Chen and Wenjun Li and Penghui Qi and Tianyu Pang and Chao Du and Wee Sun Lee and Min Lin},
      year={2025},
      eprint={2503.20783},
      archivePrefix={arXiv},
      primaryClass={cs.LG},
      url={https://arxiv.org/abs/2503.20783}, 
}

@misc{yu2025rlprextrapolatingrlvrgeneral,
      title={RLPR: Extrapolating RLVR to General Domains without Verifiers}, 
      author={Tianyu Yu and Bo Ji and Shouli Wang and Shu Yao and Zefan Wang and Ganqu Cui and Lifan Yuan and Ning Ding and Yuan Yao and Zhiyuan Liu and Maosong Sun and Tat-Seng Chua},
      year={2025},
      eprint={2506.18254},
      archivePrefix={arXiv},
      primaryClass={cs.LG},
      url={https://arxiv.org/abs/2506.18254}, 
}

@misc{zhou2025reinforcinggeneralreasoningverifiers,
      title={Reinforcing General Reasoning without Verifiers}, 
      author={Xiangxin Zhou and Zichen Liu and Anya Sims and Haonan Wang and Tianyu Pang and Chongxuan Li and Liang Wang and Min Lin and Chao Du},
      year={2025},
      eprint={2505.21493},
      archivePrefix={arXiv},
      primaryClass={cs.LG},
}

@misc{yang2025letlowprobabilitytokensoverdominate,
      title={Do Not Let Low-Probability Tokens Over-Dominate in RL for LLMs}, 
      author={Zhihe Yang and Xufang Luo and Zilong Wang and Dongqi Han and Zhiyuan He and Dongsheng Li and Yunjian Xu},
      year={2025},
      eprint={2505.12929},
      archivePrefix={arXiv},
      primaryClass={cs.CL},
}

@misc{wei2023chainofthoughtpromptingelicitsreasoning,
      title={Chain-of-Thought Prompting Elicits Reasoning in Large Language Models}, 
      author={Jason Wei and Xuezhi Wang and Dale Schuurmans and Maarten Bosma and Brian Ichter and Fei Xia and Ed Chi and Quoc Le and Denny Zhou},
      year={2023},
      eprint={2201.11903},
      archivePrefix={arXiv},
      primaryClass={cs.CL},
}

@misc{ribeiro2016whyitrustyou,
      title={"Why Should I Trust You?": Explaining the Predictions of Any Classifier}, 
      author={Marco Tulio Ribeiro and Sameer Singh and Carlos Guestrin},
      year={2016},
      eprint={1602.04938},
      archivePrefix={arXiv},
      primaryClass={cs.LG},
}

@misc{simonyan2014deepinsideconvolutionalnetworks,
      title={Deep Inside Convolutional Networks: Visualising Image Classification Models and Saliency Maps}, 
      author={Karen Simonyan and Andrea Vedaldi and Andrew Zisserman},
      year={2014},
      eprint={1312.6034},
      archivePrefix={arXiv},
      primaryClass={cs.CV},
}

@InProceedings{pmlr-v48-gal16,
  title = 	 {Dropout as a Bayesian Approximation: Representing Model Uncertainty in Deep Learning},
  author = 	 {Gal, Yarin and Ghahramani, Zoubin},
  booktitle = 	 {Proceedings of The 33rd International Conference on Machine Learning},
  pages = 	 {1050--1059},
  year = 	 {2016},
  editor = 	 {Balcan, Maria Florina and Weinberger, Kilian Q.},
  volume = 	 {48},
  series = 	 {Proceedings of Machine Learning Research},
  address = 	 {New York, New York, USA},
  month = 	 {20--22 Jun},
  publisher =    {PMLR},
  url = 	 {https://proceedings.mlr.press/v48/gal16.html}
}

@misc{jain2019attentionexplanation,
      title={Attention is not Explanation}, 
      author={Sarthak Jain and Byron C. Wallace},
      year={2019},
      eprint={1902.10186},
      archivePrefix={arXiv},
      primaryClass={cs.CL},
      url={https://arxiv.org/abs/1902.10186}, 
}

@misc{sun2025ktaemodelfreealgorithmkeytokens,
      title={KTAE: A Model-Free Algorithm to Key-Tokens Advantage Estimation in Mathematical Reasoning}, 
      author={Wei Sun and Wen Yang and Pu Jian and Qianlong Du and Fuwei Cui and Shuo Ren and Jiajun Zhang},
      year={2025},
      eprint={2505.16826},
      archivePrefix={arXiv},
      primaryClass={cs.AI},
      url={https://arxiv.org/abs/2505.16826}, 
}

\appendix
\clearpage
\section{Appendix}
\label{sec:appendix}
\begin{figure*}[t]
    \centering
    \includegraphics[width=0.99\textwidth]{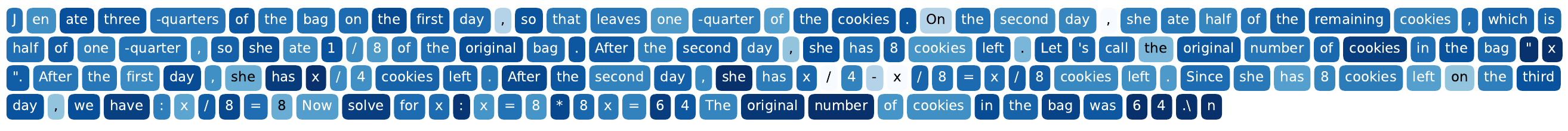}
    \includegraphics[width=\textwidth]{figs/oracle_new.pdf}
    \includegraphics[width=0.99\textwidth]{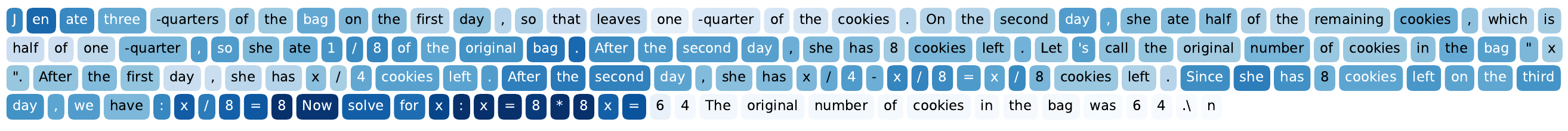}
    \caption{Token-importance visualization on a reasoning trace: OAR-G (top) vs.\ Oracle causal mask (Middle) vs.\ Entropy-Based(Bottom)}
    \label{fig:qualitative_map_2}
\end{figure*}

\subsection{Theoretical Analysis}
\label{subsec:theory}


\subsubsection{Variance from Reward Alising in GRPO}
\label{subsubsec:variance_from_grpo}
Assume there exists a latent decomposition of the sequence advantage into token-level contributions $a_t$, such that $A^{(i)} = \sum_{\tau=1}^T a_\tau$, where $a_t$ can be viewed as an unobserved per-token outcome contribution used only for analysis. The gradient contribution for a specific token at step $t$ in GRPO can be rewritten as:
\begin{equation}
    \label{eq:grad_decomposition}
    \hat{g}_t 
    = \underbrace{a_t \nabla_\theta \log \pi_t}_{\text{Token-Specific Signal}} 
    + \underbrace{\left( \sum_{\tau \neq t} a_\tau \right) \nabla_\theta \log \pi_t}_{\text{Aliased Credit (Noise)}},
\end{equation}
In an ideal token-credit setting, the update would depend primarily on $a_t$. However, GRPO effectively injects a term $\sum_{\tau \neq t} a_\tau$ into the update for token $t$, whose variance can be much larger than that of $a_t$ in long-horizon sequences:
\begin{equation}
    \mathrm{Var}\left( A^{(i)} \right) \gg \mathrm{Var}(a_t).
\end{equation}
Consequently, for tokens with low causal influence, the optimizer receives a high-variance signal largely determined by other tokens' actions. This misalignment hinders the policy from distinguishing critical reasoning steps from syntactic fillers.

\subsubsection{Why Additive Proxy Shaping may Induce Sequence-Dependent Update Scales}
\label{subsubsec:entropy_limitation}

Additive proxy shaping (e.g., entropy bonuses) is applied after GRPO's group-relative normalization, but is typically not re-normalized at the sequence level. Consequently, the effective update scale can vary across samples as a function of sequence length and how often the proxy is activated.

Define the (non-negative) proxy bonus and shaped advantage:
\begin{equation}
\begin{split}
\delta_t
=\min\!\left(\alpha \cdot \mathrm{sg}(H_t), \frac{|A|}{\kappa}\right), \\
\qquad \delta_t \ge 0, 
\qquad
\tilde{A}_t = A + \delta_t,
\label{eq:entropy_bonus_app}
\end{split}
\end{equation}
where $A$ is the GRPO group-normalized sequence advantage. The unclipped per-sequence policy gradient can be decomposed as
\begin{equation}
\begin{split}
g_{\mathrm{entropy}}
&=
\sum_{t=1}^{T} (A+\delta_t)\nabla_\theta \log \pi_\theta(y_t\mid x,y_{<t}) \\
&=
g_{\mathrm{GRPO}}
+
\sum_{t=1}^{T}\delta_t\nabla_\theta \log \pi_\theta(y_t\mid x,y_{<t}).
\end{split}
\label{eq:entropy_grad_decomp_app}
\end{equation}
The additional term scales with the bonus mass $M=\sum_{t=1}^{T}\delta_t$, which grows with sequence length $T$ and the density of high-proxy tokens. Thus, even when two samples share the same normalized $A$, their gradients can have systematically different magnitudes, effectively inducing a sample-dependent step size.

In PPO-style updates, changing the effective step size shifts the distribution of importance ratios and the realized KL to the reference policy, which may harm stability unless one applies sequence-level renormalization or re-tunes hyperparameters.

\subsubsection{Outcome-Grounded Token Importance Identification}
\label{subsubsec:cfkl}

\paragraph{Definition.}
Given a prompt $x$ and a generated sequence $y_{1:T}$, we define an \emph{outcome probe} $Z$ as the model's predictive distribution at an outcome-bearing position (e.g., the terminal next-token distribution or the start of an \texttt{<answer>} block):
\[
p \;:=\; p_\theta(\cdot \mid x, y_{1:T}).
\]
Let $y^{(-t)}$ denote a counterfactual sequence where token $y_t$ is masked (e.g., replaced by \texttt{[PAD]}), and let
\[
q_t \;:=\; p_\theta(\cdot \mid x, y^{(-t)}).
\]
We measure the influence of token $y_t$ by the KL divergence between the factual and counterfactual outcome distributions:
\begin{equation}
I_t(y) \;:=\; D_{\mathrm{KL}}(p \,\|\, q_t).
\label{eq:cfkl}
\end{equation}

\paragraph{Interpretation.}
$I_t(y)$ is \emph{outcome-grounded}: it directly quantifies how much removing $y_t$ changes the model's predicted outcome distribution, rather than relying on intrinsic signals such as token entropy.
A useful sanity connection to RLVR is that, for any event $S$ over outcomes (e.g., the verifier-accepted set of answers),
\begin{equation}
\begin{split}
\big|p(S)-q_t(S)\big|
\;\le\;\mathrm{TV}(p,q_t) \\
\;\le\;\sqrt{\tfrac12 D_{\mathrm{KL}}(p\|q_t)}
\;=\;\sqrt{\tfrac12 I_t(y)},
\label{eq:pinsker}
\end{split}
\end{equation}
where the second inequality is Pinsker's inequality. Thus, if $I_t(y)\approx 0$, masking $y_t$ cannot substantially change the probability mass of \emph{any} verifier-defined correct set, suggesting $y_t$ is counterfactually non-influential for the final outcome.

\subsubsection{Gradient Noise Suppression}
\label{subsubsec:var}

Standard GRPO broadcasts a trajectory-level advantage $\hat{A}$ to all tokens. For a nuisance token that causally contributes nothing to the reward, this broadcasting injects variance into the gradient estimate, as the token is updated based on a global signal it did not influence. We show how OAR mitigates this.

\paragraph{Claim (Noise Suppression)}
Consider a token position $k$ that is counterfactually non-influential, i.e., $I_k(y)=0$. 
Under OAR, the contribution of this position to the gradient variance is strictly suppressed compared to standard GRPO.

\textit{Proof Sketch.}
Let $s_k := \nabla_\theta \log \pi_\theta(y_k \mid x, y_{<k})$ be the score function at position $k$.
In standard GRPO, the gradient estimator for this token is $\hat{g}_{\textsc{grpo}}^{(k)} = \hat{A} \cdot s_k$.
Even if token $y_k$ is irrelevant to the task (nuisance), $\hat{A}$ varies across trajectories due to rewards earned by other tokens, and $s_k$ is non-zero (as the model still predicts the token itself). This results in gradient noise: $\mathbb{E}[\|\hat{g}_{\textsc{grpo}}^{(k)}\|^2] > 0$.

In OAR, the update is reweighted by a scalar $\omega(\hat{I}_k)$. Since $I_k(y)=0$, our weighting scheme assigns $\omega(\hat{I}_k) \le \varepsilon$ for some small $\varepsilon \ll 1$ (representing the suppression floor).
The squared norm of the OAR gradient estimator becomes:
\begin{equation}
\begin{aligned}
\mathbb{E}\big[\|\hat{g}_{\textsc{oar}}^{(k)}\|^2\big]
&= \mathbb{E}\big[\| \omega(\hat{I}_k) \cdot \hat{A} \cdot s_k \|^2\big] \\
&\le \varepsilon^2 \cdot \mathbb{E}\big[\| \hat{A} \cdot s_k \|^2\big] \\
&= \varepsilon^2 \cdot \mathbb{E}\big[\|\hat{g}_{\textsc{grpo}}^{(k)}\|^2\big].
\end{aligned}
\end{equation}
As $\varepsilon \to 0$, the gradient noise at position $k$ vanishes. Thus, OAR acts as a \emph{causal filter}, effectively freezing parameters associated with nuisance tokens and focusing the optimization on tokens that causally affect the outcome.

\subsection{Additional Token-Importance Visualizations}
\label{sec:appendix:token_vis}
This section provides additional qualitative evidence that the token-importance signals used for advantage reshaping are aligned with outcome-critical reasoning steps.Figure~\ref{fig:qualitative_map_2} shows a representative reasoning trace and compares OAR-G with an oracle causal mask and an entropy-based baseline.

\subsection{Credit Concentration Metrics}
\label{app:credit_metrics}

We quantify how concentrated token-level credit becomes during training using two auxiliary metrics computed on each sequence with a valid-token mask.

\paragraph{ESS ratio}
Let $w_t \ge 0$ denote the token weight assigned by OAR, and let $T$ be the number of valid (non-padding) tokens. Define
\begin{equation}
\begin{gathered}
\mathrm{ESS}(w)=\frac{\left(\sum_{t=1}^{T} w_t\right)^2}{\sum_{t=1}^{T} w_t^2},\\
\mathrm{ESS\text{-}ratio}(w)=\frac{\mathrm{ESS}(w)}{T}
=\frac{\left(\sum_{t=1}^{T} w_t\right)^2}{T\sum_{t=1}^{T} w_t^2}.
\end{gathered}
\end{equation}

$\mathrm{ESS\text{-}ratio}\in(0,1]$ measures the uniformity of weights: $1$ corresponds to uniform assignment, and smaller values indicate more concentrated credit.

\paragraph{Top-10\% advantage mass}
Let $A_t$ denote the token-level advantage used in the policy-gradient update. We measure what fraction of the total absolute advantage mass is carried by the top 10\% tokens (by $|A_t|$):
\begin{equation}
\begin{aligned}
m_{\mathrm{top}}=\frac{\sum_{t \in \mathrm{TopK}(|A|)} |A_t|}{\sum_{t=1}^{T} |A_t|},\\
K=\max\left(1,\left\lfloor 0.1 T \right\rfloor\right).
\end{aligned}
\end{equation}

where $\mathrm{TopK}(|A|)$ returns the indices of the $K$ largest $|A_t|$ among valid tokens. Larger $m_{\mathrm{top}}$ indicates that the update is dominated by a small subset of tokens.

\end{document}